\documentclass{article}

\usepackage{microtype}
\usepackage{graphicx}
\usepackage{subcaption}
\usepackage{booktabs} 
\usepackage{xspace}

\usepackage{hyperref}
\usepackage[table]{xcolor}
\newcommand{\best}[1]{\cellcolor{red!25}{#1}}
\newcommand{\second}[1]{\cellcolor{red!10}{#1}}

\usepackage[preprint]{icml2026}
\usepackage[utf8]{inputenc}
\usepackage[T1]{fontenc}

\usepackage{amsmath}
\usepackage{amssymb}
\usepackage{mathtools}
\usepackage{amsthm}

\usepackage[capitalize,noabbrev]{cleveref}

\usepackage[T1]{fontenc}
\usepackage{multirow}

\theoremstyle{plain}

\theoremstyle{definition}

\theoremstyle{remark}

\usepackage{pifont}
\usepackage{xcolor}
\newcommand{\cmark}{\textcolor{green!70!black}{\ding{51}}} 
\newcommand{\xmark}{\textcolor{red!70!black}{\ding{55}}}   

\usepackage{enumitem}

\makeatletter
\DeclareRobustCommand\onedot{\futurelet\@let@token\@onedot}
\def\@onedot{\ifx\@let@token.\else.\null\fi\xspace}

\def\eg{\emph{e.g}\onedot}

\makeatother

\usepackage[textsize=tiny]{todonotes}

\def\our{AnyStyle}

\usepackage{placeins}

\icmltitlerunning{AnyStyle}

\def \customparskip {0em}
\renewcommand{\paragraph}[1]{\vspace{\customparskip}\noindent\textbf{#1}}

\begin{document}

\twocolumn[
  \icmltitle{AnyStyle: Single-Pass Multimodal Stylization for 3D Gaussian Splatting}



  \icmlsetsymbol{equal}{*}

   \begin{icmlauthorlist}
    \icmlauthor{Joanna Kaleta}{uw,sano}
    \icmlauthor{Bartosz \'Swirta}{uw}
    \icmlauthor{Kacper Kania}{uw}
    \icmlauthor{Tomasz Trzcinski}{uw,ideas}
    \icmlauthor{Przemys\l{}aw Spurek}{uj,ideas}
    \icmlauthor{Marek Kowalski}{mi}
  \end{icmlauthorlist}

  \icmlaffiliation{uw}{Warsaw University of Technology}
  \icmlaffiliation{sano}{Sano Centre for Computational Medicine}
  \icmlaffiliation{uj}{Jagiellonian University}
  \icmlaffiliation{ideas}{IDEAS NCBR}
  \icmlaffiliation{mi}{Microsoft}
  
  \icmlcorrespondingauthor{}{joanna.kaleta.dokt@pw.edu.pl}

  \icmlkeywords{3D scene reconstruction, feed forward reconstruction, style transfer}

  \vskip 0.3in
]




\printAffiliationsAndNotice{}  

\begin{abstract}
The growing demand for rapid and scalable 3D asset creation has driven interest in feed-forward 3D reconstruction methods, with 3D Gaussian Splatting (3DGS) emerging as an effective scene representation. While recent approaches have demonstrated pose-free reconstruction from unposed image collections, integrating stylization or appearance control into such pipelines remains underexplored. Existing attempts largely rely on image-based conditioning, which limits both controllability and flexibility.
In this work, we introduce \our{}, a feed-forward 3D reconstruction and stylization framework that enables pose-free, zero-shot stylization through multimodal conditioning. Our method supports both textual and visual style inputs, allowing users to control the scene appearance using natural language descriptions or reference images. We propose a modular stylization architecture that requires only minimal architectural modifications and can be integrated into existing feed-forward 3D reconstruction backbones. Experiments demonstrate that \our{} improves style controllability over prior feed-forward stylization methods while preserving high-quality geometric reconstruction. A user study further confirms that \our{} achieves superior stylization quality compared to an existing state-of-the-art approach. Repository: \url{https://github.com/joaxkal/AnyStyle}.
\end{abstract}

\section{Introduction}
The rapid progress of feed-forward 3D reconstruction methods~\cite{wang2023dust3r,leroy2024grounding,ye2024no,wang2025vggt,jiang2025anysplat}, has enabled the creation of realistic 3D assets from sparse image inputs in seconds, making them increasingly relevant for applications in teleconferencing, gaming, and film production. Building upon this foundation, AnySplat introduced a pose-free reconstruction pipeline that generalizes across unconstrained image collections, representing a significant step toward scalable, data-driven 3D reconstruction.
\begin{figure}
  \centering
  \includegraphics[width=\linewidth]{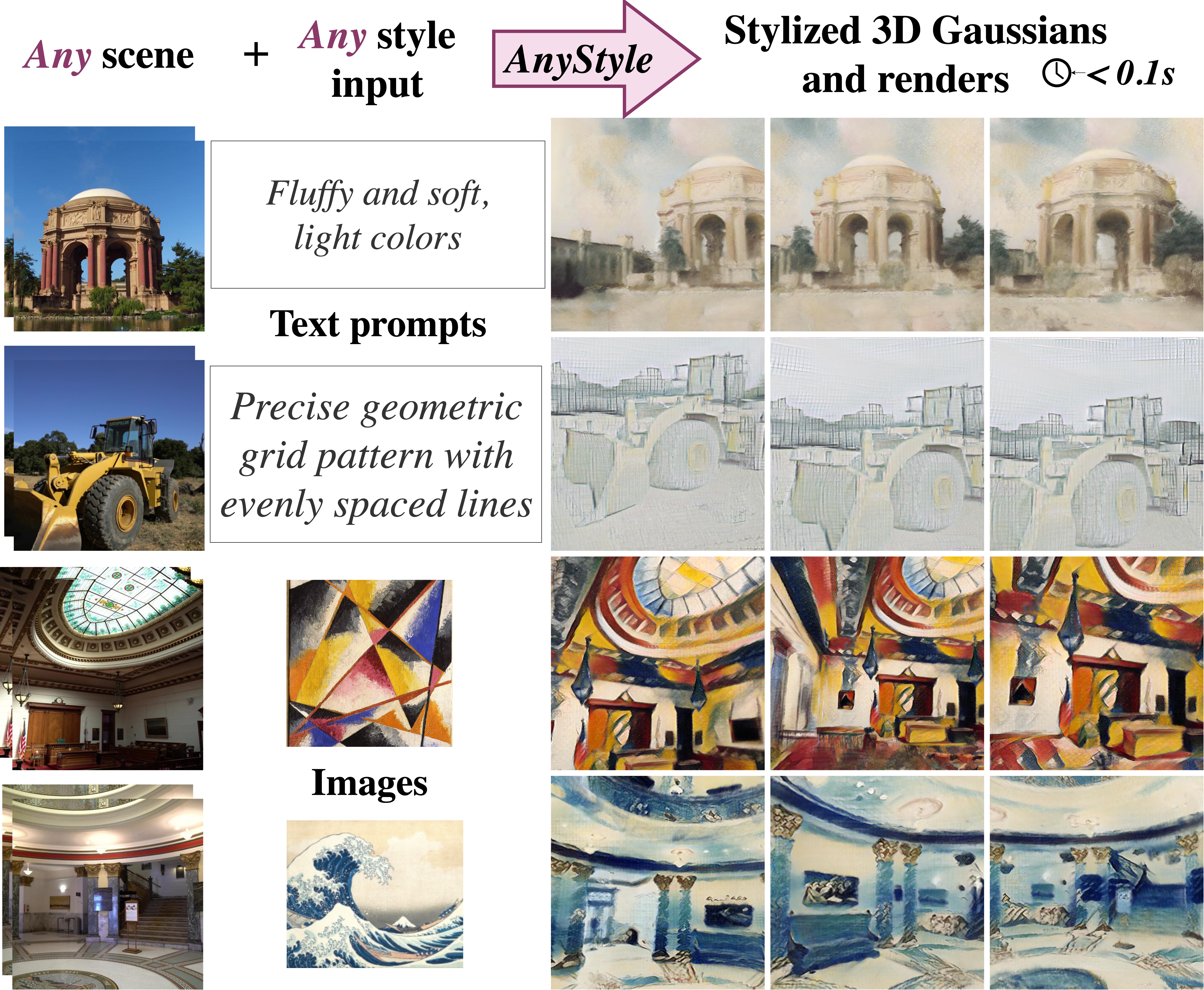}
  \caption{\textbf{Teaser.} Given a set of unposed input images and a style conditioning signal provided as either text or an image, our method generates a stylized 3D scene represented with  3D Gaussian Splats in a single forward pass. The reconstructed scene can be stylized in under 0.1 second per input content image.}
  \label{fig:teaser}
  \vspace{-2.05em}
\end{figure}

More recently, Styl3R~\cite{wang2025styl3r} and Stylos~\cite{liu2025stylos} have explored simultaneous 3D reconstruction and artistic stylization, enabling users to transfer the visual style of a reference image directly into the reconstructed 3D scene. These models demonstrated that stylization can be integrated into a single forward pass, eliminating the need for iterative optimization required by earlier works \cite{howil2025clipgaussian,liu2024stylegaussian,kovacs2024G}. 

However, these approaches remain limited in two aspects. First, the style control is restricted to visual exemplars: a single style image determines the overall aesthetic, leaving little room for fine-grained or semantic manipulation. Second, existing 3D stylization approaches \cite{wang2025styl3r, liu2025stylos} use architectural components that require training the underlying 3D reconstruction method from scratch. This design makes the application of those methods to new architectures difficult and reduces modularity.

We introduce \our{}, a multimodal, zero-shot stylization framework for pose-free 3D Gaussian Splatting that overcomes those limitations. \our{} unifies textual and visual style control through a shared CLIP-based embedding space, enabling stylization to be guided by style images, text prompts, or continuous interpolation between the two. This multimodal conditioning provides both semantic flexibility and fine-grained control: for instance, when a reference image produces a nearly desired appearance, user can refine the result by steering the style embedding toward textual cues such as ``expressive brushwork'' or ``softer color tones,'' without the need to search for an alternative style image.

Crucially, our approach enables practical and efficient stylization through its architecture-agnostic design. Unlike existing feed-forward methods \cite{liu2025stylos, wang2025styl3r} that require training both geometric and stylization branches jointly from scratch, our method initializes from a pretrained 3D reconstruction model and only adds a lightweight style injection module that is fine-tuned alongside the backbone. Our style injection approach is inspired by ControlNet \cite{Zhang_2023_ICCVcontrolnet}, which introduced zero-initialized convolutions for conditional control in 2D diffusion models. To the best of our knowledge, we are the first to adapt this mechanism for stylization in feed-forward 3D reconstruction.

Unlike attention-based approaches \cite{chung2024style_trainingfree, styleid, Han_2025_ICCVstylebooth} that modify internal Query, Key, and Value computations, our method injects style through additive conditioning via zero-initialized convolutions. This design offers two key advantages: \textbf{(1) modularity} -- the style injection module can be added to any pretrained attention-based model without altering its internal mechanics; and \textbf{(2) architectural generality} -- stylization is decoupled from attention mechanism, enabling style control at any stage of the network, including tokens that do not re-enter attention layers. This flexibility allows integration into diverse transformer-based reconstruction models regardless of their attention patterns or layer configurations. We further validate this property in our ablation study by stylizing only tokens routed directly to prediction head.

Experiments demonstrate that AnyStyle achieves superior stylization quality compared to existing methods, with the best ArtFID scores among feed-forward approaches and statistically significant preference in a user study. Our method maintains this quality advantage while not requiring training from scratch.

\begin{figure*}
  \centering
  \includegraphics[width=0.95\textwidth]{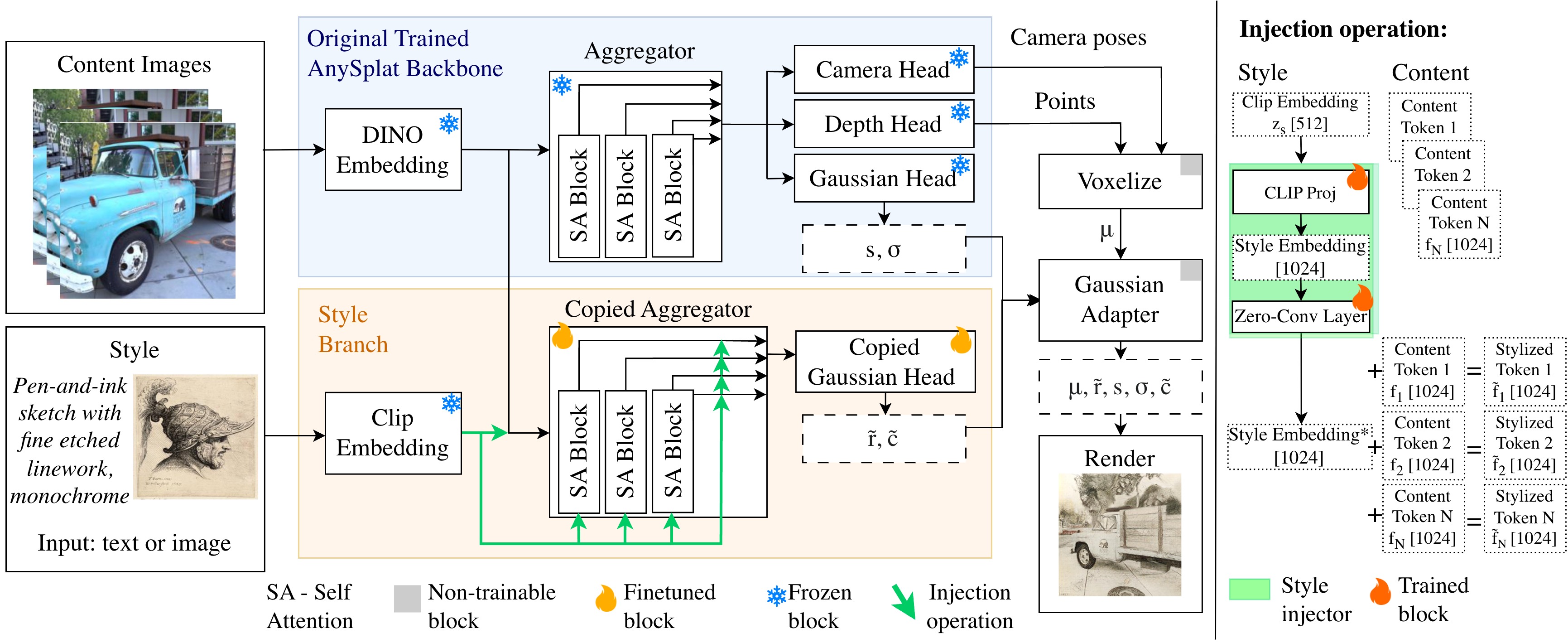}
  \caption{\textbf{Method overview.} AnyStyle takes unposed content images of a scene together with an arbitrary style input (text or image) and produces a stylized 3D Gaussian representation from which novel stylized views can be rendered. The architecture follows a dual-branch design that decouples geometric reconstruction from appearance stylization. Content images are processed by a pretrained frozen backbone to recover geometry and camera poses, while the style signal is embedded using CLIP and applied only within the style branch to control scene appearance. A pretrained AnySplat model initializes the copied Aggregator $\tilde{A}$ and Gaussian Head $\tilde{H}_{gs}$ in the style branch. These components are subsequently fine-tuned with CLIP-based conditioning via style injection. The outputs of the two branches are combined by a Gaussian Adapter and rendered to produce the final stylized views.}
  \label{fig:method}
  \vspace{-0.5cm}
\end{figure*}

In summary, AnyStyle advances controllable 3D stylization by offering:
\vspace{-1em}
\begin{itemize}[leftmargin=*]
    \item Novel architecture applying zero-convolution conditioning to style injection and enabling multimodal style control via CLIP embeddings that combine text and image inputs,
    \item Semantic and continuous style refinement through prompt manipulation or interpolation,
    \item Architecture-agnostic design, enabling integration into any transformer-based model without modifying the original architecture or requiring additional pretraining.
\end{itemize}
\vspace{-1em}
These contributions enable expressive, flexible, and intuitive control over 3D stylization in a single feed-forward pass.

\section{Related Work}



\begin{table}[t]
\centering
\caption{Comparison of properties of discussed 3D reconstruction and stylization methods.}
\resizebox{0.48\textwidth}{!}{
\begin{tabular}{@{}l@{\;}ccccc@{}}
\toprule
\multirow{2}{*}{Method} &
Multimodal &
Scene &
Style &
Pose &
Fast \\
& Style Input
& Zero-Shot
& Zero-Shot
& Free
& Inference \\
\midrule
ARF, SNeRF, StylizedNeRF 
& \xmark & \xmark & \xmark & \xmark & \xmark \\
StyleRF
& \xmark & \xmark & \cmark & \xmark & \xmark \\
SGSST, G-Style, StylizedGS
& \xmark & \xmark & \xmark & \xmark & \cmark \\
StyleGaussian
& \xmark & \xmark & \cmark & \xmark & \cmark \\
ClipGaussian
& \cmark & \xmark & \xmark & \xmark & \cmark \\
Stylos, Styl3r
& \xmark & \cmark & \cmark & \cmark & \cmark \\
\midrule
\textbf{\our{}}
& \cmark & \cmark & \cmark & \cmark & \cmark \\
\bottomrule
\end{tabular}
}
\label{tab:method_comparison}
\vspace{-1em}
\end{table}

\textbf{Neural 3D Scene Representations.} Neural Radiance Fields (NeRFs)~\cite{mildenhall2021nerf} revolutionized 3D scene reconstruction by representing scenes as continuous volumetric functions optimized from multi-view images. While achieving photorealistic novel view synthesis, NeRFs are difficult to integrate with classical graphics engines and many variants are slow to render.

3D Gaussian Splatting (3DGS)~\cite{kerbl20233d} addresses these limitations by representing scenes as collections of 3D Gaussians with learnable parameters, enabling real-time rendering while maintaining high visual quality. This explicit representation has become the foundation for numerous downstream applications, including style transfer.

\textbf{Feed-Forward 3D Reconstruction.}
Traditional NeRF and 3DGS methods require per-scene optimization, taking tens of minutes on a single GPU to reconstruct each scene. This computational burden limits their practical applicability, particularly for applications requiring real-time or near-real-time processing. To address this limitation, recent methods leverage learned structural priors~\cite{wang2023dust3r,leroy2024grounding,wang2025vggt,charatan2024pixelsplat,chen2024mvsplat,szymanowicz2024splatter} to reconstruct 3D point maps in a single feed-forward pass. Building on these principles, subsequent approaches directly predict parameters of 3D Gaussians~\cite{smart2024splatt3r,ye2024no}, eliminating the iterative optimization process entirely. Most notably, AnySplat~\cite{jiang2025anysplat} extends VGGT~\cite{wang2025vggt} to reconstruct 3D Gaussians without reliance on calibrated cameras, enabling reconstruction from casual image captures. The feed-forward nature of these methods is particularly advantageous for 3D stylization, as it enables applying artistic styles to novel scenes without scene-specific fine-tuning. We leverage AnySplat's architecture as the foundation for our stylization framework.

\textbf{Style Transfer: From 2D to 3D.}
 Classical 2D style transfer methods~\cite{gatys2016image} match statistics of VGG-extracted features (\eg, Gram matrices) between content and style images. This approach enabled the development of faster feed-forward methods~\cite{huang2017arbitrary, li2017universal,li2019learning,park2019arbitrary,liu2021adaattn,wu2021styleformer,deng2022stytr2,zhang2024s2wat,tang2023master}, and eventually text-guided domain adaptation \cite{gal2022stylegan} and stylization~\cite{kwon2022clipstyler, radford2021learning, suresh2024fastclipstyler} using CLIP for fine-grained control through natural language. However, naïve per-view stylization of 3D scenes produces severe multi-view inconsistencies.

Early 3D methods trained NeRF models on stylized multi-view images~\cite{zhang2022arf, nguyenphuoc2022snerf, huang2022stylizednerf, liu2023stylerf}, requiring lengthy per-scene optimization. The emergence of 3DGS enabled faster approaches~\cite{liu2023stylegaussian, galerne2025sgsst, kovacs2024gstyle, zhang2024stylizedgs, yu2024instantstylegaussian}, with some incorporating text prompts~\cite{howil2025clipgaussian}. However, all these methods rely on per-scene optimization, requiring scene-specific reconstruction and, in most cases, additional stylization optimization. An overview comparing 3D style transfer methods is shown in \cref{tab:method_comparison}.

Recent feed-forward alternatives~\cite{wang2025styl3r, liu2025stylos} eliminate per-scene optimization by predicting stylized 3D Gaussians in a single pass. However, \textit{they rely solely on a single style image as input}, which limits fine-grained control over specific stylistic elements (\eg, color palette vs. brush strokes). Moreover, they require heavy training process in which both, geometric backbone and stylization branch are trained jointly.

\our{} addresses these limitations through multi-modal conditioning with both style images and text prompts, using  CLIP-based style embedding, enabling zero-shot style transfer in a single feed-forward pass with fine-grained artistic control. Inspired by ControlNet, our approach allows for extending any attention-based pre-trained model with stylization branch. 




\section{Method}
\paragraph{AnySplat Backbone.}
\label{sec:anysplat_backbone}
\our{} is constructed upon AnySplat~\cite{jiang2025anysplat}, a feed-forward 3D reconstruction framework that predicts a complete 3D Gaussian scene representation from unposed multi-view images in a single forward pass. Given a set of input images $\{\mathbf{I}_i\}_{i=1}^N$, each image is processed by a frozen DINO \cite{caron2021emerging_dino} feature extractor $f(\cdot)$ to produce local patch features: $\mathbf{F}_i = f(\mathbf{I}_i)$ for $i = 1, \dots, N$.

The Aggregator $A$ is composed of multiple self-attention blocks with alternating local and global self-attention and processes the DINO patch features of the input. At each transformer layer $l \in \{1,\dots,L\}$, the Aggregator produces a sequence of token representations, denoted as $\mathbf{T}^{(l)}$. Token representations from a selected subset of intermediate layers $\mathcal{S} \subset \{1,\dots,L-1\}$, together with the final-layer tokens $\mathbf{T}^{(L)}$, are used to form a unified scene representation, which we denote compactly as
\begin{equation}
\mathbf{T}_{\mathrm{agg}} = \{\mathbf{T}^{(l)} \mid l \in \mathcal{S} \cup \{L\}\} = A(\{\mathbf{F}_i\}_{i=1}^N).
\end{equation}
This merged token representation is then provided as input to three specialized reconstruction heads. The Camera Head $H_{\text{cam}}$ estimates camera parameters for each input view. The Depth Head $H_{\text{depth}}$ predicts depth and unprojects it to a set of 3D point locations $\{\boldsymbol{\mu}_i\}_{i=1}^{n}$ that define the centers of the Gaussian primitives for each pixel in each input view. The Gaussian Head $H_{\text{gs}}$ predicts the remaining geometric and appearance attributes, yielding a set of $n$ anisotropic 3D Gaussian primitives representing scene geometry and appearance:
\begin{equation}
\mathcal{G}_i =
\left(
\mathbf{\mu}_i, \mathbf{r}_i, \mathbf{s}_i,
\sigma_i,
\mathbf{c}_i
\right),
\end{equation}
where $\mathbf{r}_i$ denotes the rotation, $\mathbf{s}_i$ the scale, $\sigma_i$ the opacity, and $\mathbf{c}_i$ the spherical-harmonic color coefficients.

To reduce redundancy while preserving geometric and appearance fidelity, AnySplat employs a voxelization module that clusters nearby Gaussians in 3D space and merges them via confidence-weighted averaging. The resulting Gaussian set is rendered using differentiable Gaussian splatting to produce high-quality novel views.

\begin{figure*}
  \centering
  \includegraphics[width=\linewidth]{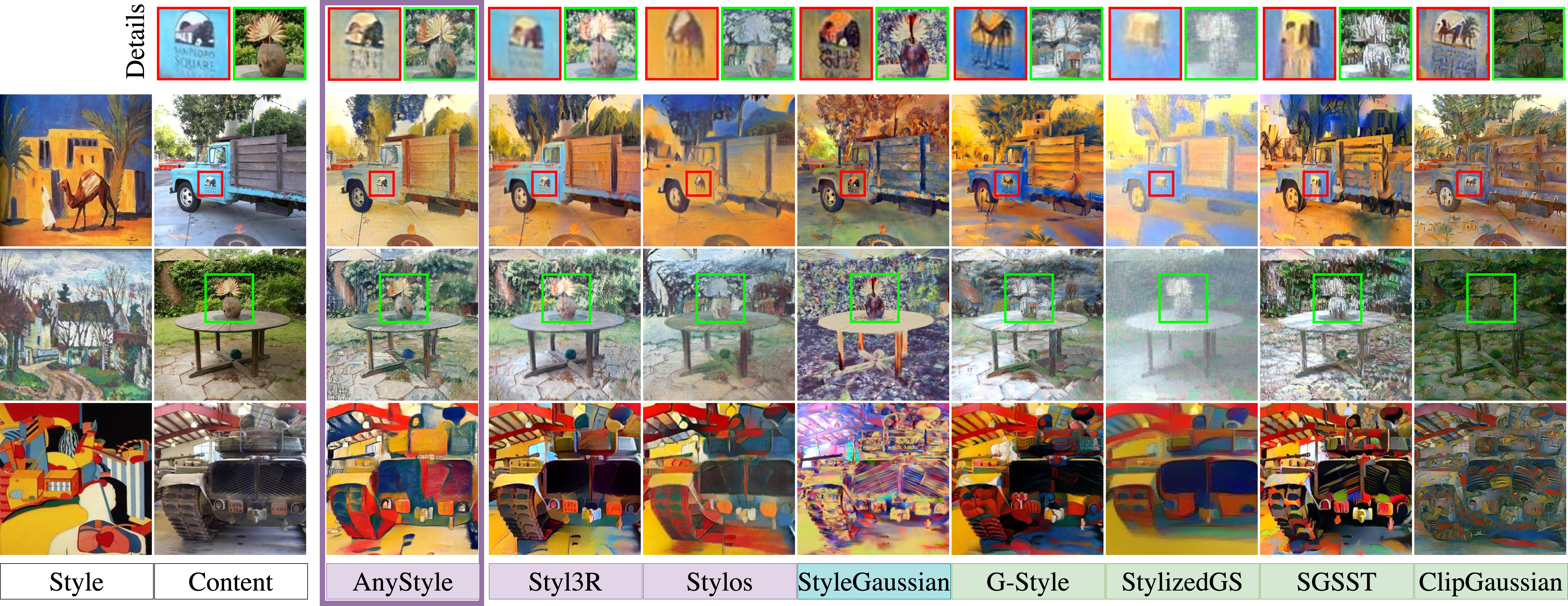}
  \caption{\textbf{Comparison between AnyStyle and existing 3D style transfer methods} with different architectural designs: feed-forward (purple), per-scene optimization (green), and hybrid approaches (blue). Our method achieves high-quality style transfer while faithfully preserving fine details (top row) as well as overall scene structure. All compared methods are conditioned on a \textbf{style image}.}
  \label{fig:sota-comparison}
  \vspace{-0.5cm}
\end{figure*}

\paragraph{\our{} Architecture Overview.}
\label{sec:architecture_overview}
The \our{} architecture adopts a dual-branch design that explicitly decouples geometric reconstruction from artistic stylization. As illustrated in \cref{fig:method}, the framework consists of a \textit{Frozen Backbone} original weights and a dedicated \textit{Style Branch}. The backbone remains frozen throughout training in order to preserve the structural integrity of the reconstructed scene, including camera poses and the underlying 3D point distribution. We use AnySplat as  backbone, but thanks to \our{}'s architecture we could switch to a different backbone, e.g. DepthAnything 3 \cite{depthanything3}. 

In parallel to the frozen backbone, the \textit{Style Branch} comprises a copied Gaussian head $\tilde{H}_{gs}$ and a copied Aggregator $\tilde{A}$ module (see \textbf{Injection in Style Branch} paragraph in~\cref{style_inj_sec}), which are fine-tuned to adapt the appearance of the scene. Stylization is driven by multimodal conditioning, where a style input---either text or an image---is embedded into a shared CLIP feature space. The core mechanism enabling appearance adaptation is the \textit{style injection} operation, which is implemented via a dedicated \textit{Style Injector} module.

Finally, a \textit{Gaussian Adapter} combines the geometric parameters $\mathbf{\mu}_i, \mathbf{s}_i, \sigma_i$ obtained from the frozen backbone with the stylized features $\tilde{\mathbf{r}}_i, \tilde{\mathbf{c}}_i$ produced by the Style Branch. These features are predicted by the Style Branch to improve expressiveness without degrading geometric accuracy (e.g., by shrinking Gaussians or making them overly transparent). The resulting set of stylized 3D Gaussian primitives is rendered using a differentiable renderer to produce the final output views.

\paragraph{Multimodal Style Conditioning.}
To support both text- and image-based style control, we embed all style inputs into a shared CLIP feature space. Specifically, we employ Long-CLIP~\cite{zhang2024longclip}, which improves the modeling of long-range semantics and fine-grained stylistic attributes, enabling richer and more expressive style representations from text.

Given a style image $\mathbf{S}_I$ or a text prompt $\mathbf{S}_T$, we extract a Long-CLIP embedding
\begin{equation}
\mathbf{z}_s = E_{\text{CLIP}}(\mathbf{S}_I)
\;\;\text{or}\;\;
\mathbf{z}_s = E_{\text{CLIP}}(\mathbf{S}_T),
\quad \mathbf{z}_s \in \mathbb{R}^{d_s} .
\end{equation}
Although Long-CLIP aligns text and image embeddings within a shared latent space, a residual modality gap persists in practice. To mitigate this effect, we alternate between text-conditioned and image-conditioned training batches. This training strategy encourages the network to learn a unified style representation that generalizes across both modalities.

\paragraph{Zero-Convolution Style Injection.}
\label{sec:zero_conv}
To enable appearance stylization, we inject style information into feature representations processed by the network. Style conditioning can be applied at selected intermediate layers within the Aggregator or/and to the features at the output of the Aggregator. 

For this purpose, we introduce a \textbf{Style Injector} module, denoted as $\mathcal{I}(\cdot, \cdot)$. For each style input (text or image), we extract the style embedding $\mathbf{z}_s$ and project it to the token dimension using a lightweight multi-layer perceptron (MLP). The resulting projected style embedding is then injected into the features via additive modulation. 

Naively injecting style features directly into a representation can destabilize training
, 
\cite{Zhang_2023_ICCVcontrolnet}. To address this issue, we adopt zero-initialized convolutional adapters, inspired by \cite{Zhang_2023_ICCVcontrolnet, lu2025align3r}. At initialization, these adapters output zeros, ensuring that the network behaves identically to the pretrained AnySplat model before stylization training begins.

Formally, given a token $\mathbf{f} \in \mathbb{R}^{d_f}$ and a style embedding $\mathbf{z}_s \in \mathbb{R}^{d_s}$, the Style Injector produces a style-conditioned representation defined as
\begin{equation}
\tilde{\mathbf{f}} = \mathcal{I}(\mathbf{f}, \mathbf{z}_s)
= \mathbf{f} + \mathrm{ZeroConv}(\mathrm{Proj}(\mathbf{z}_s)),
\end{equation}
where $\mathrm{Proj}: \mathbb{R}^{d_s} \rightarrow \mathbb{R}^{d_f}$ denotes a projection MLP and $\mathrm{ZeroConv}(\cdot)$ is a zero-initialized convolutional layer as in \cite{Zhang_2023_ICCVcontrolnet}. $\mathrm{ZeroConv}(\cdot)$ is a single layer, with $\mathrm{1x1}$ convolution, zero padding and $\mathrm{ch_{in} = ch_{out}}$, that performs weighted multiplication of input channels. 

In \our{}, each injection location corresponds to a specific Aggregator layer and employs its own dedicated Style Injector instance. This allows layer-wise adaptation to feature representations of different dimensionalities.

\paragraph{Style Injection in Style Branch.}
\label{style_inj_sec}
In this paragraph we consider two  location at which the Style Injector $\mathcal{I}(\cdot,\cdot)$ is applied. 

\textbf{(i) Head injection.}
Style injection is applied to the tokens \textbf{exiting} $\tilde{A}$, included in representation $T_{agg}$, that is fed into the Gaussian Head:
\begin{equation}
\tilde{\mathbf{t}}^{(l)} = \mathcal{I}^{(l)}(\mathbf{t}^{(l)}, \mathbf{z}_s),
\qquad \forall \mathbf{t}^{(l)} \in \mathbf{T}^{(l)}, \; \mathbf{T}^{(l)} \in \mathbf{T}_{\mathrm{agg}}.
\end{equation}
\textbf{(ii) Aggregator injection}
Style information is injected into token representations at selected intermediate layers, which are processed further by the aggregator $\tilde{A}$.

Let $\mathbf{T_{x}}^{(l)}$ denote the set of token representations produced at $\tilde{A}$ layer $l$. We denote by $\mathcal{P} \subset \{1,\dots,L-1\}$ the set of selected intermediate layers at which expressive style injection is applied. For each layer $l \in \mathcal{P}$, style injection is applied to all tokens at that layer:
\begin{equation}
\tilde{\mathbf{t}}^{(l)}_{x} = \mathcal{I}^{(l)}(\mathbf{t}^{(l)}_{x}, \mathbf{z}_s),
\qquad \forall \mathbf{t}^{(l)}_{x} \in \mathbf{T}^{(l)}_x, \; l \in \mathcal{P}.
\end{equation}
This allows style cues to influence token interactions within the $\tilde{A}$ itself.

\paragraph{Training losses.}
Our training objective builds on losses used in feed-forward 3D stylization and augments them with CLIP-based alignment. While prior methods mainly rely on VGG\cite{guan2019deep_vgg} perceptual content and style losses, we find that combining them with a CLIP directional loss provides complementary supervision: VGG losses preserve structure and texture statistics, whereas CLIP encourages semantic style alignment. Together, they enable stronger stylistic transfer than either objective alone.

\textbf{Content Loss.}
To preserve scene structure and avoid over-stylization, we employ a content loss computed between the rendered stylized input views $\hat{\mathbf{I}}$ and the original RGB input views $\mathbf{I}$ in a deep feature space. Let $\phi(\cdot)$ denote a pretrained VGG19 network, and let $\phi^{(l)}(\cdot)$ denote the activation at layer $l$. The content loss is defined as
\begin{equation}
\mathcal{L}_{\text{content}} =
\sum_{l \in \mathcal{K}_{\text{content}}}
\left\|
\phi^{(l)}(\hat{\mathbf{I}})
-
\phi^{(l)}(\mathbf{I})
\right\|_2^2 ,
\end{equation}
where $\mathcal{K}_{\text{content}} = \{\texttt{relu3\_1}, \texttt{relu4\_1}\}$ denotes the set of VGG layers used for content supervision. Following classical neural style transfer, deeper VGG layers are used for content supervision to preserve high-level structure, while multiple shallow-to-deep layers capture multi-scale texture statistics for style~\cite{huang2017arbitrary, zhang2022arf}. In line with Styl3R~\cite{wang2025styl3r}, we use both \texttt{relu3\_1} and \texttt{relu4\_1} for the content loss, which empirically better preserves scene structure than single-layer supervision commonly used in earlier style transfer works~\cite{zhang2022arf, galerne2025sgsst}.

\textbf{Style Loss.}
We adopt a style loss that matches channel-wise feature statistics between the rendered image $\hat{\mathbf{I}}$ and a reference style image $\mathbf{S}_I$. Style features are extracted from the same $\phi$ at multiple layers. The style loss is defined as
\begin{equation}
\begin{aligned}
\mathcal{L}_{\text{style}} =
\sum_{l \in \mathcal{K}_{\text{style}}}
\Big(
\big\|
\mu\!\left(\phi^{(l)}(\hat{\mathbf{I}})\right)
-
\mu\!\left(\phi^{(l)}(\mathbf{S}_I)\right)
\big\|_2^2 \\
+
\big\|
\sigma\!\left(\phi^{(l)}(\hat{\mathbf{I}})\right)
-
\sigma\!\left(\phi^{(l)}(\mathbf{S}_I)\right)
\big\|_2^2
\Big),
\end{aligned}
\end{equation}
where $\mu(\cdot)$ and $\sigma(\cdot)$ denote the channel-wise mean and standard deviation, and
$\mathcal{K}_{\text{style}} = \{\texttt{relu1\_1}, \texttt{relu2\_1}, \texttt{relu3\_1}, \texttt{relu4\_1}\}$.

\begin{table}
\centering
\caption{Stylization quality, as measured by ArtScore and ArtFID (abbreviated as Score and FID respectively), and stylization time comparisons with recent 3D stylization models. Results for per-scene methods were imported from \cite{liu2025stylos}. \textit{img} and \textit{txt} denote inference with style image and style text prompt respectively. We mark with red the best and second best scores.}
\label{tab:artscore}
\resizebox{0.49\textwidth}{!}{
\begin{tabular}{lccccccccc}
\toprule
\multirow{2}[2]{*}{Method}
& \multicolumn{2}{c}{Train}
& \multicolumn{2}{c}{Truck}
& \multicolumn{2}{c}{M60}
& \multicolumn{2}{c}{Garden}
& \multirow{2}[2]{*}{Time$\downarrow$} \\
\cmidrule(lr){2-3} \cmidrule(lr){4-5} \cmidrule(lr){6-7} \cmidrule(lr){8-9}
& Score$\uparrow$ & FID$\downarrow$
& Score$\uparrow$ & FID$\downarrow$
& Score$\uparrow$ & FID$\downarrow$
& Score$\uparrow$ & FID$\downarrow$
&  \\
\midrule
StyleGaussian  & 0.78 & 52.79 & 5.76 & 44.93 & 8.63 & 47.48 & 9.38 & 41.14 & 165 m \\
G-Style  & 9.52 & 23.24 & 9.67 & \best{22.15} & 9.73 & \best{22.36} & 8.98 & 25.76 & 14.5 m\\
StylizedGS  & 4.95 & 40.79 & 6.45 & 42.07 & 7.39 & 53.24 & 6.99 & 65.15 & 13.3 m \\
SGSST & 1.84 & 38.24 & 5.34 & 32.34 & 5.26 & 38.73 & 4.89 & 33.54 & 35.0 m \\
\midrule
Styl3R & 4.67 & 27.74 & 7.07 & 27.28 & 8.43 & 23.98 & 8.61 & 29.73 & 0.16 s\\
Stylos  & \second{9.50} & 26.40 & 9.70 & 28.71 & 9.37 & 27.44 & 9.34 & 28.06 & 0.05 s \\
\midrule
\our{}$_{img}$ & 8.84 & \best{22.86} & 9.59 & \second{22.95} & 9.47 & \second{22.81} & 9.48 & \best{22.32} & 0.07s\\
\our{}$_{txt}$  & \best{9.87} & 24.41 & \best{10.56} & 24.67 & \best{10.20} & 24.16 & \best{10.46} & 24.20 & 0.07s\\
\our{}$_{Head, img}$ & 8.45 & \second{22.89} & 9.97 & 25.29 & 9.03 & \second{22.81} & 9.72 & \second{23.85} & 0.05s\\
\our{}$_{Head, txt}$ & 8.69 & 24.18 & \second{10.10} & 27.89 & \second{9.54} & 24.34 & \best{10.07} & 25.56 & 0.05s \\
\bottomrule
\end{tabular}
 }
\vspace{-0.0cm}
\end{table}

\begin{table}
    \centering
    \caption{Ablation study. Stylization quality is measured by ArtScore and ArtFID (abbreviated as Score and FID respectively). We present comparisons for different versions of the \textit{Head} model.}
    \label{tab:ablation-quality}
    \resizebox{0.49\textwidth}{!}{
    \begin{tabular}{lcccccccc}
        \toprule
        \multirow{2}[2]{*}{Method}
        & \multicolumn{2}{c}{Train}
        & \multicolumn{2}{c}{Truck}
        & \multicolumn{2}{c}{M60}
        & \multicolumn{2}{c}{Garden} \\
        \cmidrule(lr){2-3} \cmidrule(lr){4-5} \cmidrule(lr){6-7} \cmidrule(lr){8-9}
        & Score$\uparrow$ & FID$\downarrow$
        & Score$\uparrow$ & FID$\downarrow$
        & Score$\uparrow$ & FID$\downarrow$
        & Score$\uparrow$ & FID$\downarrow$ \\
        \midrule
        \multicolumn{9}{c}{ \our{}$_{Head, img}$} \\
        \midrule
        base config & 8.45 & 22.89 & 9.97 & 25.29 & 9.03 & 22.81 & 9.72 & 23.85 \\
         w/ all geom features & 8.77 & 23.08 & 9.61 & 25.07 & 8.77 & 23.30 & 9.57 & 23.65\\
         w/o CLIP losses &5.80 & 23.26 & 8.75 & 26.35 & 7.30 & 25.00 & 8.60 & 23.85\\
         w/o Style loss & 4.67 & 27.65 & 6.56 & 27.20 & 5.65 & 24.95 & 5.93 & 25.76 \\
        
        \midrule
        \multicolumn{9}{c}{ \our{}$_{Head, txt}$} \\
        \midrule
        base config & 8.69 & 24.18 & 10.10 & 27.89 & 9.54 & 24.34 & 10.07 & 25.56 \\
        w/o text in training & 9.89 & 29.40 & 10.53 & 32.59 & 9.90 & 30.04 & 10.59 & 31.00\\

        \bottomrule
    \end{tabular}
    }
    \vspace{-1em}
\end{table}
\paragraph{CLIP Loss.}
\label{method:clip-loss}
To enhance style alignment between the stylized output and the desired style, we follow the CLIP-based style transfer works \cite{kwon2022clipstyler, howil2025clipgaussian} and employ a CLIP directional loss. Let $\mathbf{z}_{\tilde{\mathbf{I}}}$ and $\mathbf{z}_{\mathbf{I}}$ denote the CLIP embeddings of the stylized rendering $\tilde{\mathbf{I}}$ and the original rendering $\mathbf{I}$, respectively. Let $\mathbf{z}_{\text{Photo}}$ denote the embedding of a universal neutral prompt \emph{``Photo''}. We define the CLIP directional loss between an original image $\mathbf{I}$, a stylized image $\hat{\mathbf{I}}$, and a style input $\mathbf{S}$ as
\begin{equation}
\mathcal{L}_{\text{CLIP}}(\mathbf{I}, \tilde{\mathbf{I}}, \mathbf{S}) =
1 - \cos\Big(
\mathbf{z}_{\tilde{\mathbf{I}}} - \mathbf{z}_{\mathbf{I}},
\;
\mathbf{z}_s - \mathbf{z}_{\text{Photo}}
\Big),
\end{equation}
In addition to the global loss, we apply the same CLIP loss to local image patches sampled from the stylized rendering. The patch-level CLIP loss enforces \emph{local} style consistency beyond global semantic alignment, which can overlook fine-grained textures. As observed in~\cite{kwon2022clipstyler}, applying directional CLIP loss to randomly sampled patches improves the transfer of spatially invariant texture cues.The patch-level CLIP loss is defined as
\begin{equation}
\mathcal{L}_{\text{CLIP}}^{\text{p}} =
\frac{1}{|\mathcal{N}_{\text{patch}}|}
\sum_{i \in \mathcal{N}_{\text{patch}}}
\mathcal{L}_{\text{CLIP}}\big(\mathbf{I}, p_i(\tilde{\mathbf{I}}), \mathbf{S}\big),
\end{equation}
where $p_i(\hat{\mathbf{I}})$ denotes the $i$-th patch, and $\mathcal{N}_{\text{patch}}$ denotes number of patches. Random perspective augmentations are applied to each patch prior to CLIP embedding extraction, encouraging fine-grained and locally consistent stylization.

We define the final objective as:
\begin{equation}
\mathcal{L} =
\lambda_c \mathcal{L}_{\text{content}}
+ \lambda_s \mathcal{L}_{\text{style}}
+ \lambda_{\text{clip}} \mathcal{L}_{\text{CLIP}}
+\lambda_{\text{clip}}^{p} \mathcal{L}_{\text{CLIP}}^p.
\end{equation}

\section{Experiments}
We provide additional results including qualitative results on scenes from other dataset, training details, code, extended analysis, user study details and limitations in the Supplementary Material. We encourage the reader to see the supplementary video, images and analysis.  

\begin{figure}
  \centering
  \includegraphics[width=0.85\linewidth]{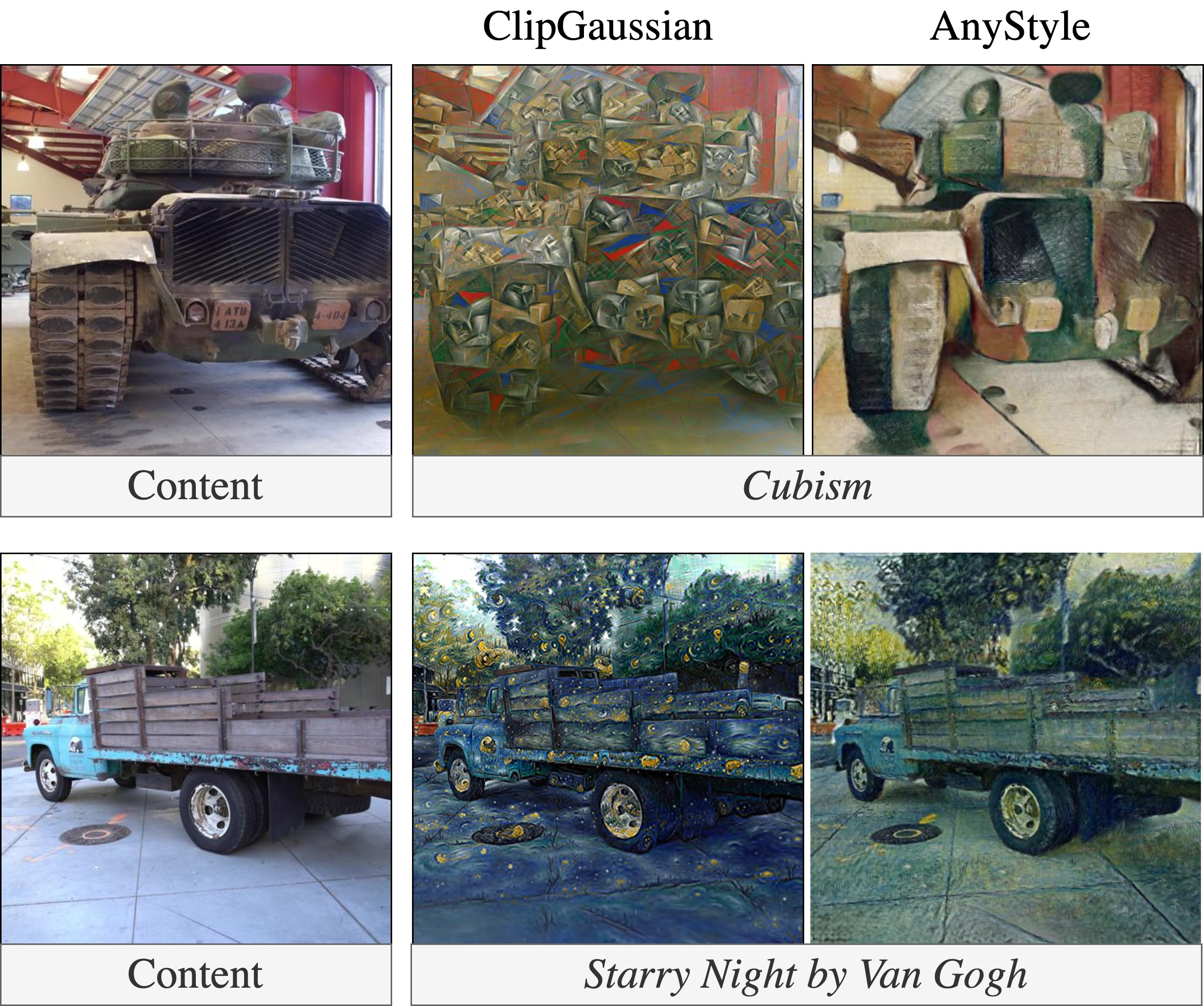}
  \caption{\textbf{Stylization using text prompts.} We compare \our{} with ClipGaussian \cite{howil2025clipgaussian}, which requires per-scene optimization (>20min). Despite using identical text prompts, ClipGaussian introduces semantic artifacts from the style input.}
  \label{fig:clipgaussian}
  \vspace{-1em}
\end{figure}
\paragraph{Datasets.}
\label{sec:datasets}
During training, for content supervision we used DL3DV-480P dataset \cite{ling2023dl3dv10k}. It provides wide range of both indoor at outdoor scenes. For style supervision we use the WikiArt dataset \cite{wikiart}, which is extensively utilized in style transfer tasks. For stylization with text input, we generated descriptions for the style images with Mini-CPM-V4.5 model \cite{yao2024minicpm}.

For evaluation, following \cite{liu2025stylos}, we used 3 scenes (\textit{Train}, \textit{Truck}, \textit{M60}) from \textit{TnT} dataset \cite{Knapitsch2017tanks}, and \textit{Garden} from \textit{Mip-NeRF360} dataset \cite{barron2022mipnerf360} as content data and 50 Wikiart style images provided by \cite{liu2025stylos}, excluded from the training. 
%
\paragraph{Baselines.}
\label{sec:baselines}
We closely follow \cite{liu2025stylos} evaluation procedure, thus we can directly compare our method to multiple other approaches, both feed-forward: Stylos \cite{liu2025stylos}, Styl3R \cite{wang2025styl3r} and based on per-scene optimization: StyleGaussian \cite{liu2023stylegaussian}, G-Style \cite{kovacs2024gstyle}, StylizedGS \cite{zhang2024stylizedgs}, SGSST \cite{galerne2025sgsst}, ClipGaussian \cite{howil2025clipgaussian}. \our{} is tested in a zero-shot manner without prior knowledge of either the scenes or styles.

\paragraph{Stylization -- quantitative evaluation.}
We evaluate stylization quality using \textbf{ArtScore}~\cite{chen2024artscore}, which measures how closely an image resembles authentic artwork, and \textbf{ArtFID}~\cite{wright_gcpr_2022}, which measures how well an image matches a particular style. We conduct evaluations under both image-based and text-based style conditioning. Quantitative comparisons with prior methods are reported in \cref{tab:artscore}.

\our{} achieves the best ArtFID across all scenes for both image and text input compared to other feed-forward methods. In terms of ArtScore, $\text{\our{}}_{txt}$ also achieves the best results across all scenes, while $\text{\our{}}_{img}$  beats most existing methods. We observe that text-conditioned models obtain higher ArtScore but slightly worse ArtFID than image-conditioned counterparts. We attribute this to the nature of the stylization guided by text prompts which inherently provides less visual cues.

\begin{figure}[tbp]
  \centering
  \includegraphics[width=\linewidth]{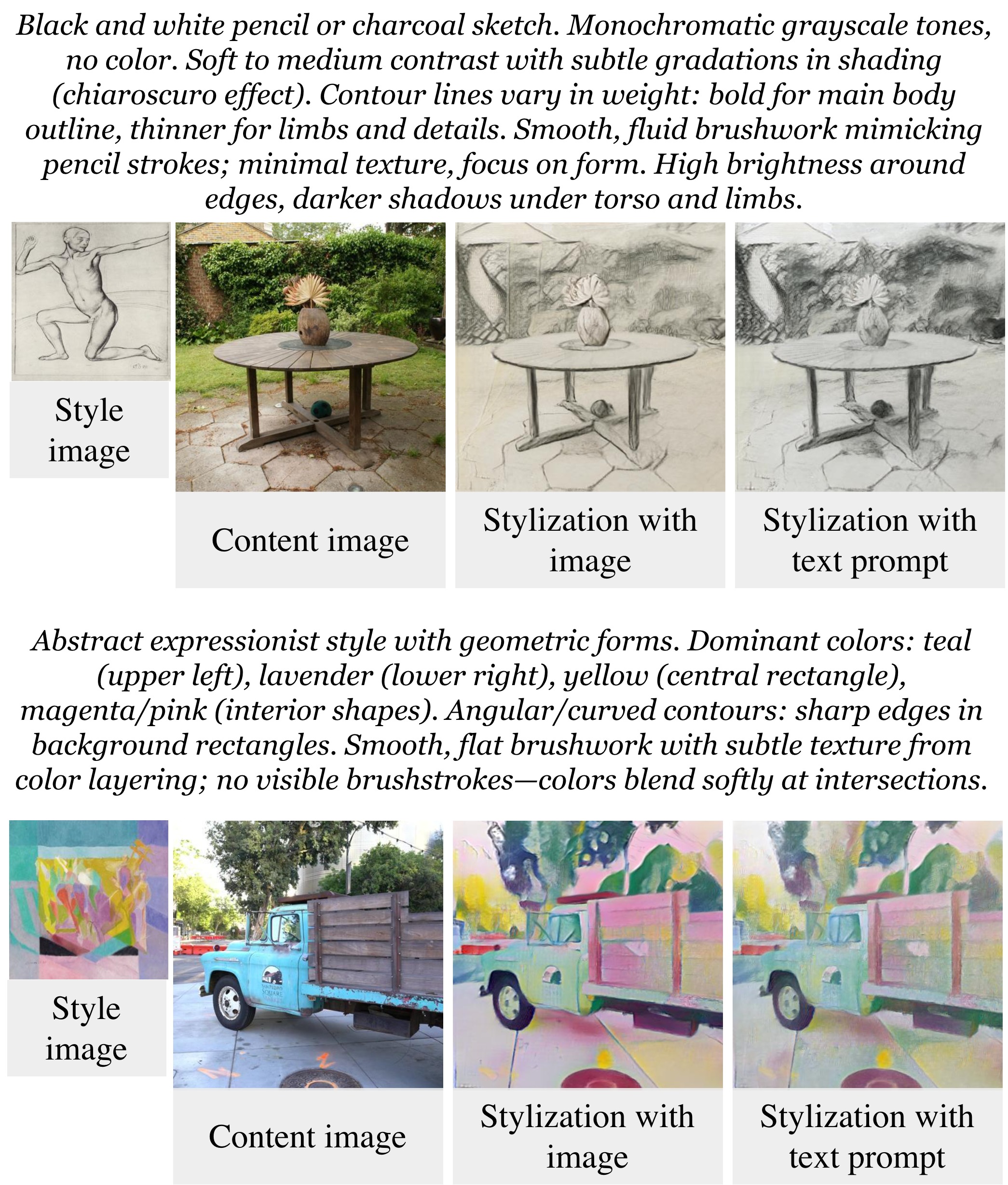}
  \caption{\textbf{Stylization with text prompts vs. images.} We compare renderings conditioned either on a reference test style image or on a textual description generated by Mini-CPM-V4.5 for that image. Our method achieves coherent and plausible stylization across both modalities. Please note that due to the inherently lower amount of information encoded in text prompts and more ambiguous nature of natural language, text-based conditioning cannot reproduce the rendered appearance exactly the same as image-based conditioning.}
  \label{fig:text_vs_img}
  \vspace{-1.5em}
\end{figure}

\begin{figure}[tbp]
  \centering
  \includegraphics[width=\linewidth]{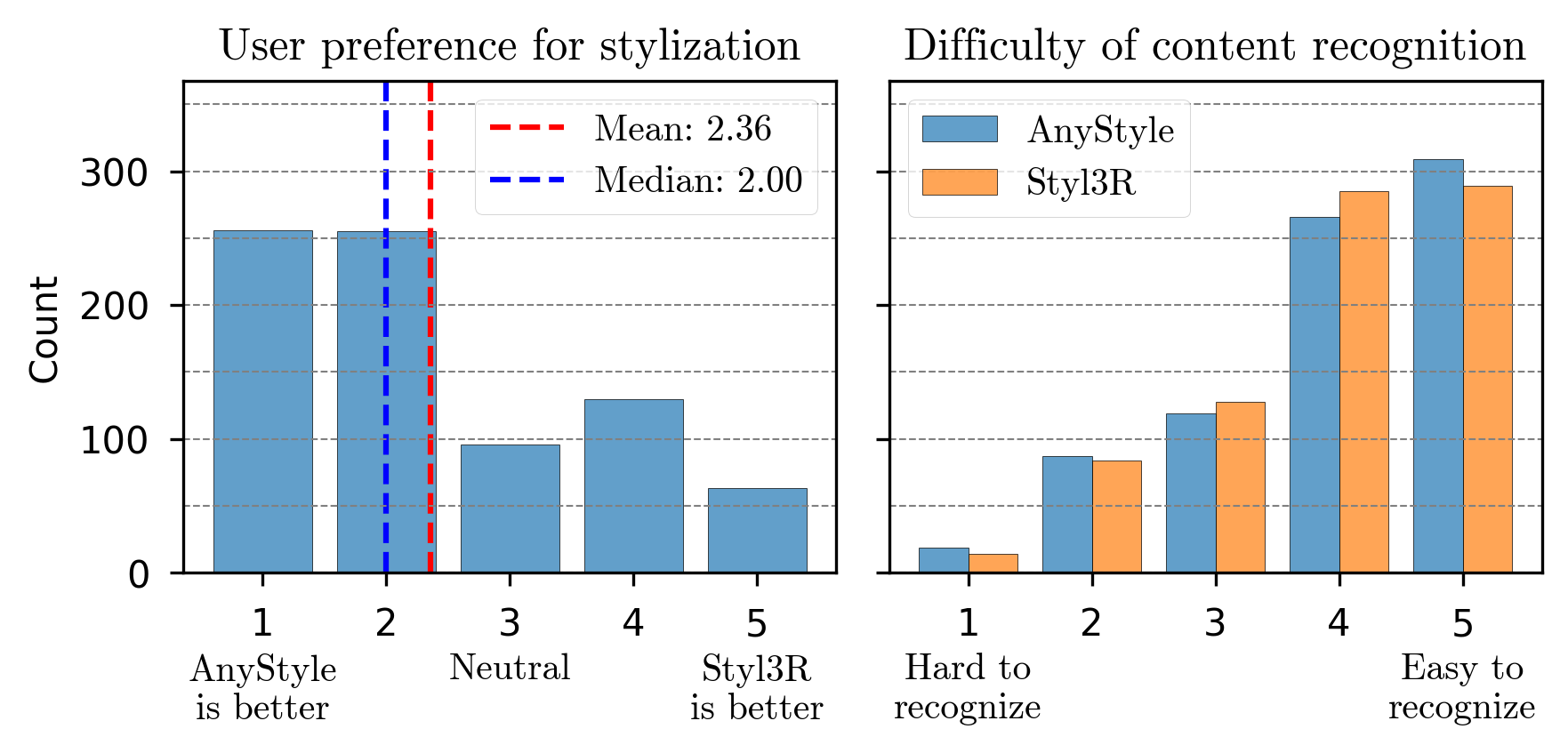}
  \caption{\textbf{User study.} According to responders,images  generated by \our{} better follow reference style. Content preservation is comparable for both methods.}
  \label{fig:user-study}
  \vspace{-1em}
\end{figure}

\paragraph{Stylization -- qualitative evaluation.}
\label{sec:user-study}
Qualitative comparisons between our method and current state-of-the-art models are illustrated in \cref{fig:sota-comparison} and \cref{fig:clipgaussian}, while \cref{fig:text_vs_img} provides a visual analysis of image- versus text-based style conditioning.

\begin{figure}[tbp]
  \centering
  \includegraphics[width=\linewidth]{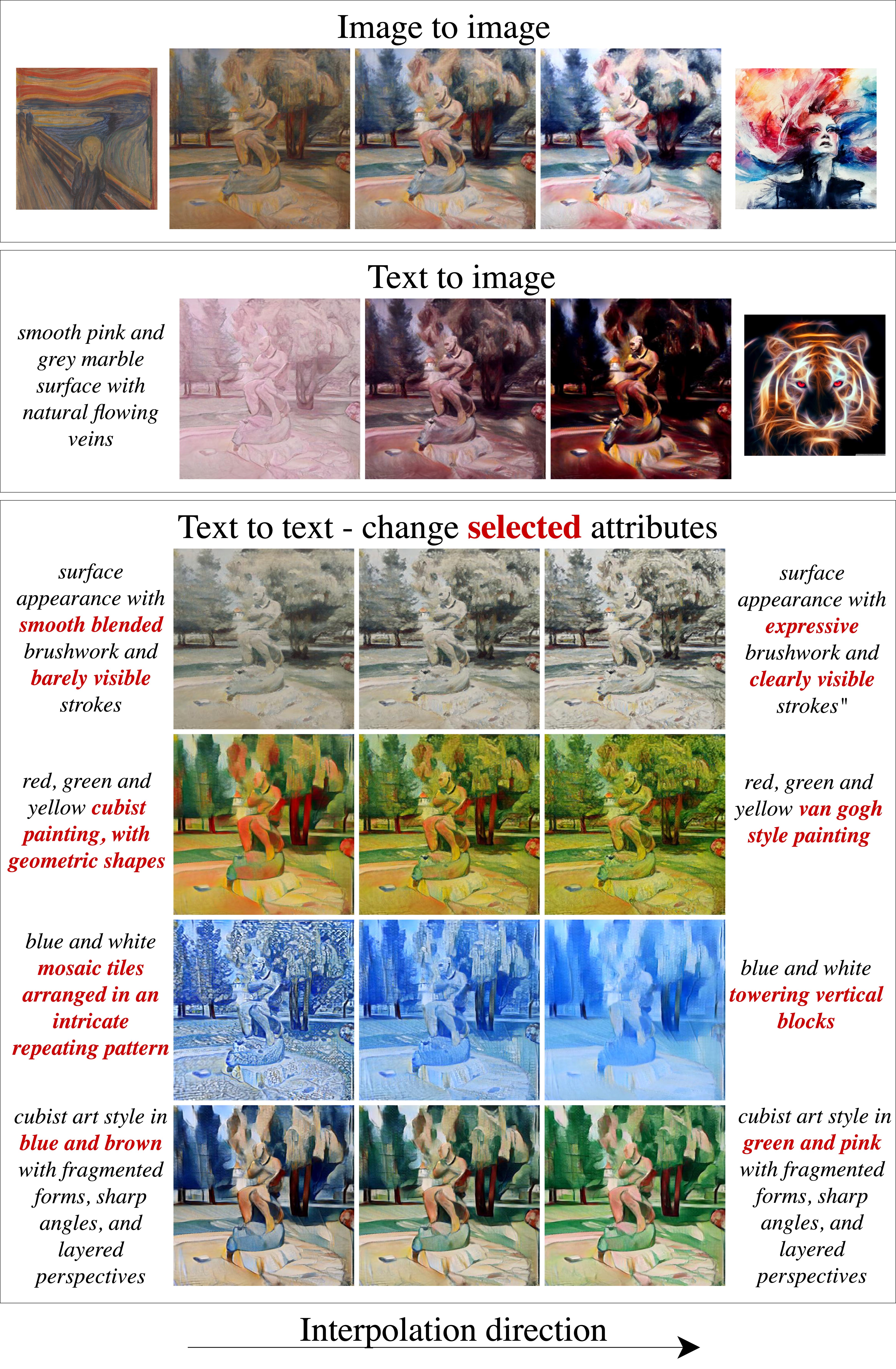}
  \caption{\textbf{Style interpolation.} Thanks to the unified CLIP embedding space, our method enables smooth interpolation between two style images, between an image and a text prompt, as well as between two text prompts. Compared to image-only style transfer methods, the latter capability provides greater control over the stylization process and supports an iterative workflow for refining specific stylistic attributes (highlighted in red).}
  \label{fig:interpolation}
  \vspace{-1.5em}
\end{figure}

\paragraph{Style interpolation.}
Our method operates in a unified CLIP-based embedding space, enabling flexible control over the stylization process. Qualitative results are shown in~\cref{fig:interpolation}. Beyond standard interpolation between two distinct styles, our model also supports interpolation between closely related style descriptions that differ only in specific attributes. This allows fine-grained adjustment of stylistic properties without requiring the user to search for a new reference image, which is typically necessary in existing image-based style transfer methods.

\begin{figure}[tbp]
  \centering
  \includegraphics[width=\linewidth]{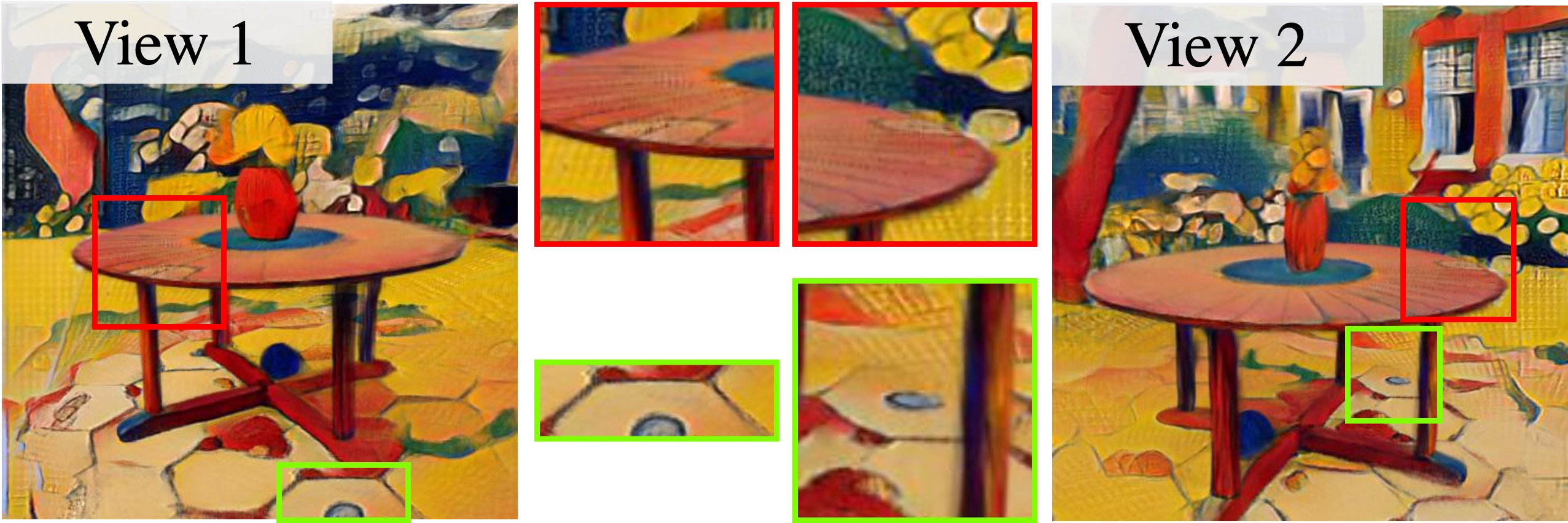}
  \caption{\textbf{Multi-view consistency of the stylization.} By directly updating 3D Gaussian representation, our method ensures multi-view consistency of stylization while preserving fine-grained details across viewpoints. Floor dot (marked in green) and table discoloration (marked in red) remain consistent across views.}
  \label{fig:consistency}
  \vspace{-1em}
\end{figure}

\paragraph{Multi-view consistency.}
\label{sec:multi-view}
We provide a qualitative evaluation of multi-view consistency in~\cref{fig:consistency}, where our method preserves a coherent appearance of scene details across viewpoints. For additional video results, quantitative evaluation and detailed discussion, we refer to~\cref{apx:mv-const} in the Supplementary Material.

To further assess perceptual quality, we conducted an online user study comparing our method with Styl3R (StylOS code was unavailable at submission time). A total of 40 participants evaluated 20 cases, each with three questions related to stylization quality and content preservation. They rated the results on a 5-point Likert scale~\cite{likert1932technique}. This resulted in 800 responses per question. The results are shown in \cref{fig:user-study}. Statistical significance was evaluated using one-sample t-tests and Wilcoxon signed-rank tests. The observed preference for our method was statistically significant; detailed test statistics are reported in the Supplementary Material.

\begin{figure}[tbp]
  \centering
  \includegraphics[width=\linewidth]{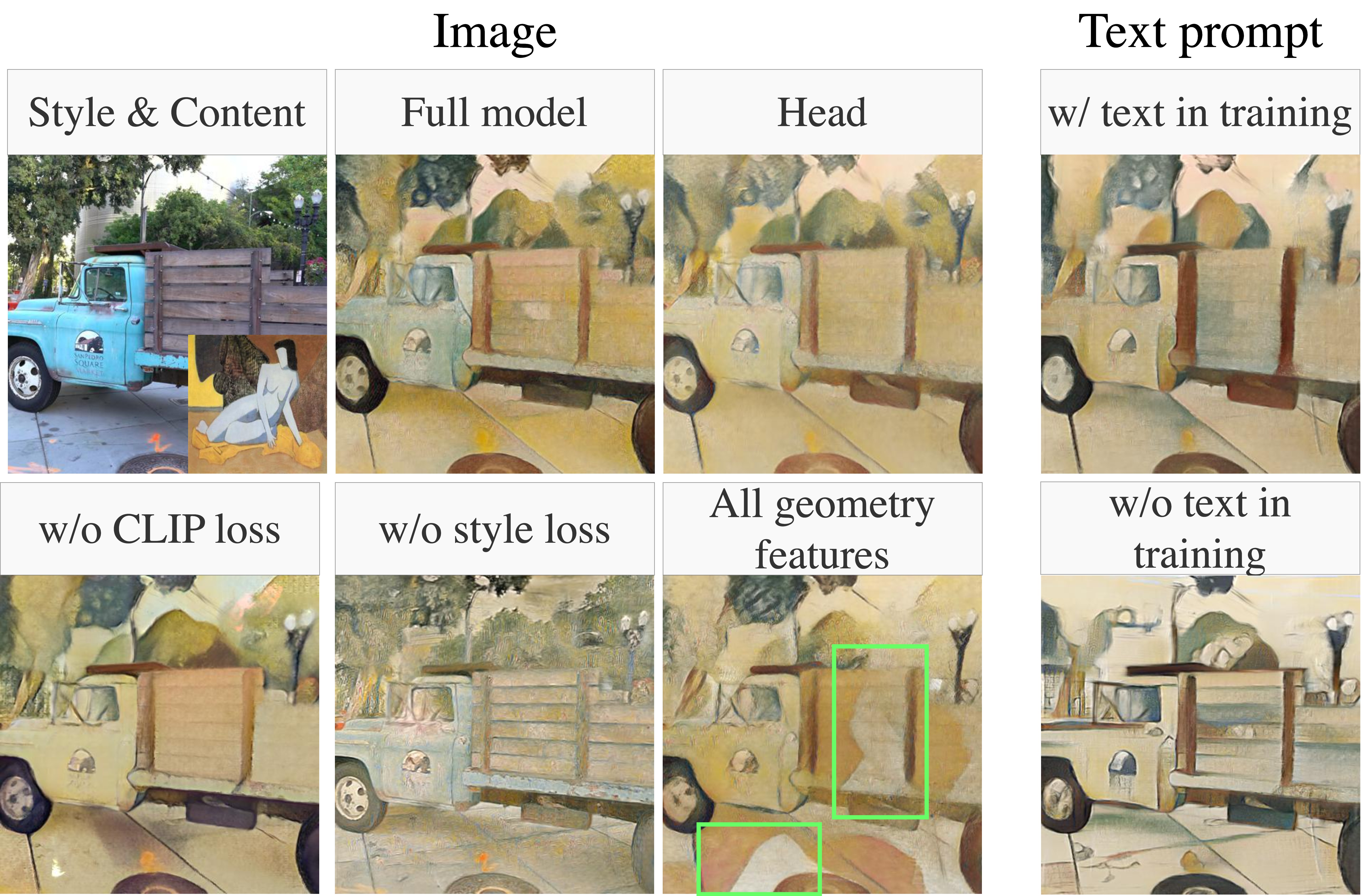}
  \caption{\textbf{Ablation study.}
The first three columns show stylization conditioned on a reference style image. Full model produces richer and more vivid colors, while the \textit{Head} variant yields slightly flatter appearances. Removing the style loss prevents accurate style learning, whereas removing the CLIP directional loss reduces color expressiveness. Finetuning all geometric features leads to visual inconsistencies (highlighted in green), such as unintended transparent artifacts mimicking cubist features. The last column shows text-conditioned stylization using a prompt corresponding to the reference style image; models trained with textual supervision follow the target style more closely than those trained without it.}
  \label{fig:ablation}
  \vspace{-2em}
\end{figure}

\paragraph{Ablation study.}
\label{sec:ablation-study}
First, we perform an ablation study on two architectural design choices: (i) the full model where style injection is applied in both the Aggregator and the Head, and both $\tilde{H}_{gs}$ and $\tilde{A}$ are fine-tuned; and (ii) \textbf{a lightweight setting, denoted as \textit{Head}}, where style injection is applied only to tokens entering $\tilde{H}_{gs}$ and only $\tilde{H}_{gs}$ is fine-tuned. As shown in \cref{tab:artscore}, both variants achieve strong results compared to existing, particularly feed-forward, methods. The \textit{Head} version typically yields slightly higher FID than full model, indicating weaker alignment with the target style images. This ablation demonstrates that our injection mechanism remains effective even when applied solely to tokens forwarded directly to prediction head.
 
We further ablate the \textit{Head} variant. The results are presented in \cref{fig:ablation} and \cref{tab:ablation-quality}. CLIP-based losses alone provide a weak supervisory signal, while style loss alone degrades color alignment; removing either leads to consistent performance drops across all metrics. Updating all Gaussian parameters, although not harmful in terms of quantitative metrics, introduces visible geometric artifacts and degrades multi-view consistency. Finally, excluding textual conditioning during training significantly degrades text-driven stylization, leading to results that poorly follow the target style. This is also reflected in a substantially higher ArtFID for the w/o text variant.

\section{Conclusions}
We introduced \our{}, a feed-forward framework for multimodal, zero-shot stylization of 3D Gaussian Splatting scenes. Unlike prior approaches that rely solely on image-based conditioning or require retraining pipelines from scratch, \our{} enables stylization via an architecture-agnostic style injection mechanism attachable to pretrained 3D reconstruction backbones with minimal modification. Operating in a shared CLIP embedding space, the method supports text- and image-driven control, as well as continuous interpolation between styles, enabling flexible and semantically meaningful manipulation of scene appearance. By decoupling geometry from appearance, \our{} preserves structural consistency while allowing expressive stylization. Our experiments demonstrate strong quantitative and perceptual results, achieving superior stylization quality among feed-forward methods.


\section{Acknowledgements}
J. Kaleta was supported by National Science Centre, Poland grant no 2022/47/O/ST6/01407. P. Spurek was supported by the project Effective Rendering of 3D Objects Using Gaussian Splatting in an Augmented Reality Environment (FENG.02.02-IP.05-0114/23), carried out under the First Team programme of the Foundation for Polish Science and co-financed by the European Union through the European Funds for Smart Economy 2021–2027 (FENG). We gratefully acknowledge Polish high-performance computing infrastructure PLGrid (HPC Center: ACK Cyfronet AGH) for providing computer facilities and support within computational grants no. PLG/2025/018551 and no. PLG/2024/017221.


\bibliographystyle{icml2026}

\newpage
\appendix
\onecolumn

\section{Supplementary overview}
This Supplementary Material is organized as follows:
\begin{itemize}
    \item \textbf{Extended qualitative results} (\cref{sec:ext-qual}): additional stylized renders on RE10K, including both image-guided and text-prompt stylization.
    \item \textbf{User study} (\cref{sec:user-study-appendix}): study design, questions, and statistical analysis of participant preferences.
    \item \textbf{Training details} (Sec.~\ref{sec:train-details}): code, datasets, optimization setup, loss weights, and architecture choices.
    \item \textbf{Multi-view consistency: Analysis and discussion} (Sec.~\ref{apx:mv-const}): consistency evaluation protocol and limitations of the warping-based metric.
    \item \textbf{Extended quantitative results} (Sec.~\ref{sec:ext-quant}): additional metrics for stylized outputs.
    \item \textbf{Limitations and failure cases} (Sec.~\ref{sec:limitations}): representative \our{} failures for image- and text-conditioned stylization.
\end{itemize}

\section{Extended qualitative results }
\label{sec:ext-qual}
\begin{figure*}[h!]
  \centering
  \includegraphics[width=0.75\linewidth]{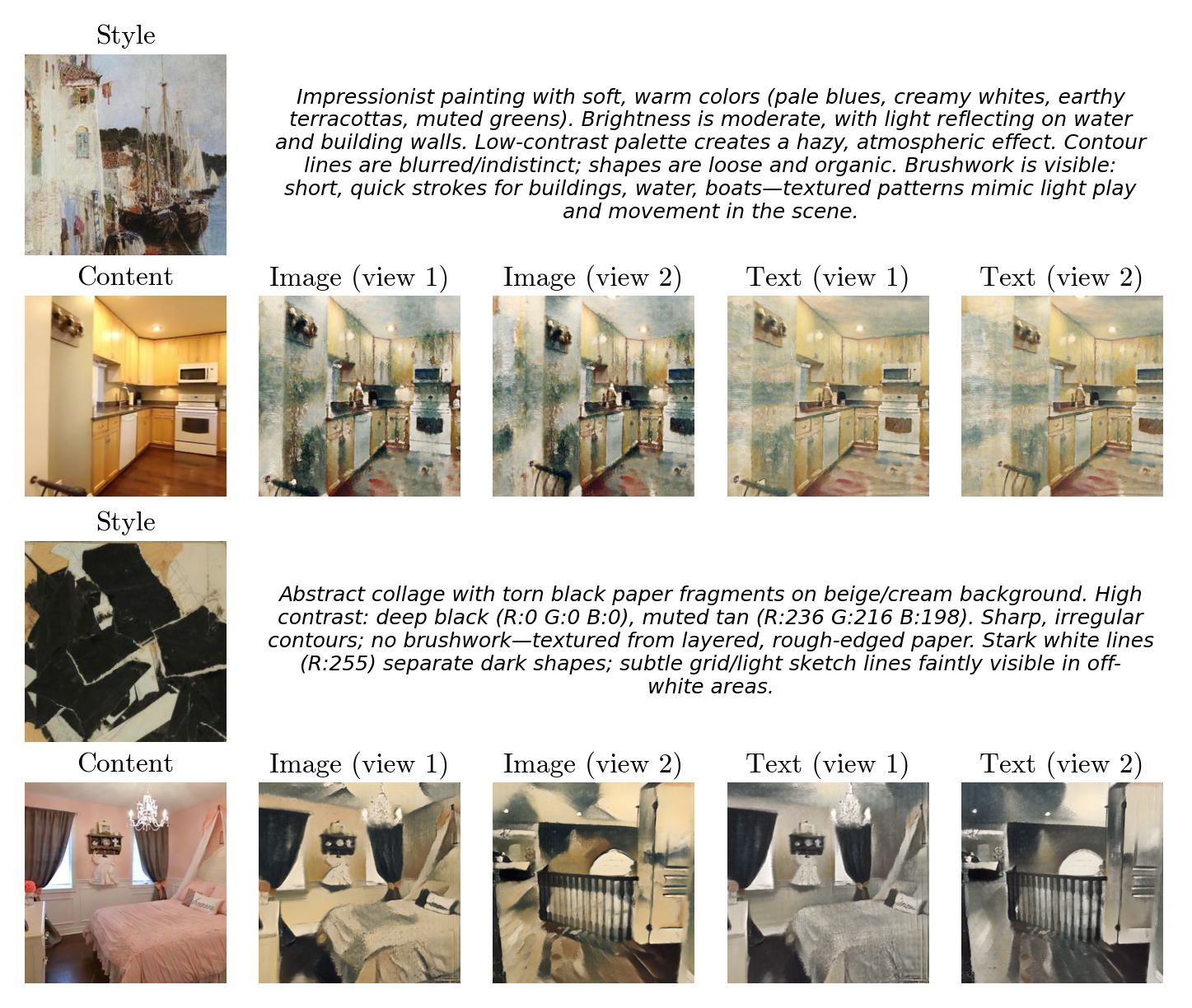}
  \caption{Additional qualitative results of our method on RE10K dataset. We present stylization with image and text prompt.}
  \label{fig:appx-qual-agg-1}
\end{figure*}

\begin{figure*}
  \centering
  \includegraphics[width=0.75\linewidth]{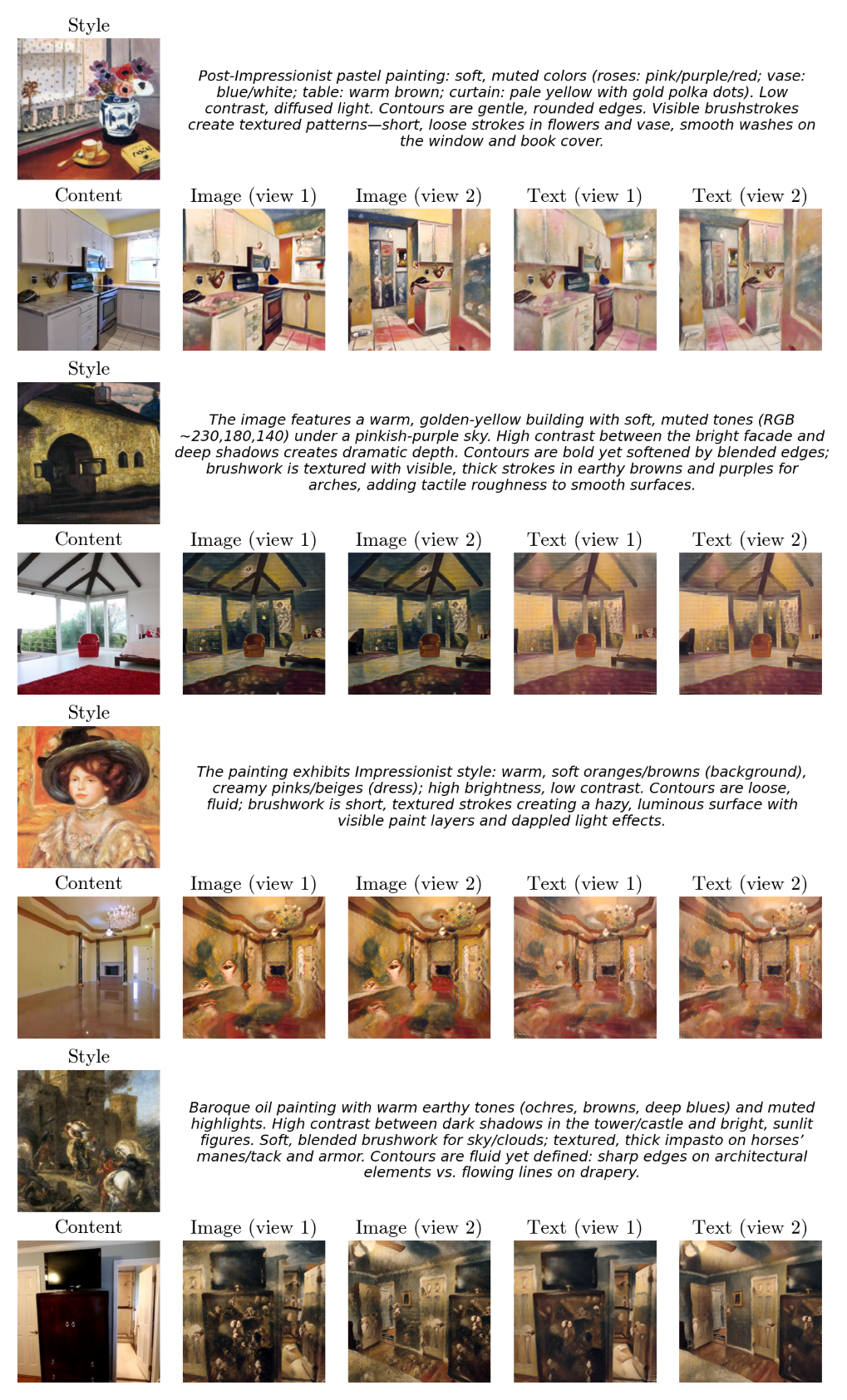}
  \caption{Additional qualitative results of our method on RE10K dataset. We present stylization with image and text prompt.}
  \label{fig:appx-qual-agg-2}
\end{figure*}

\begin{figure*}
  \centering
  \includegraphics[width=0.75\linewidth]{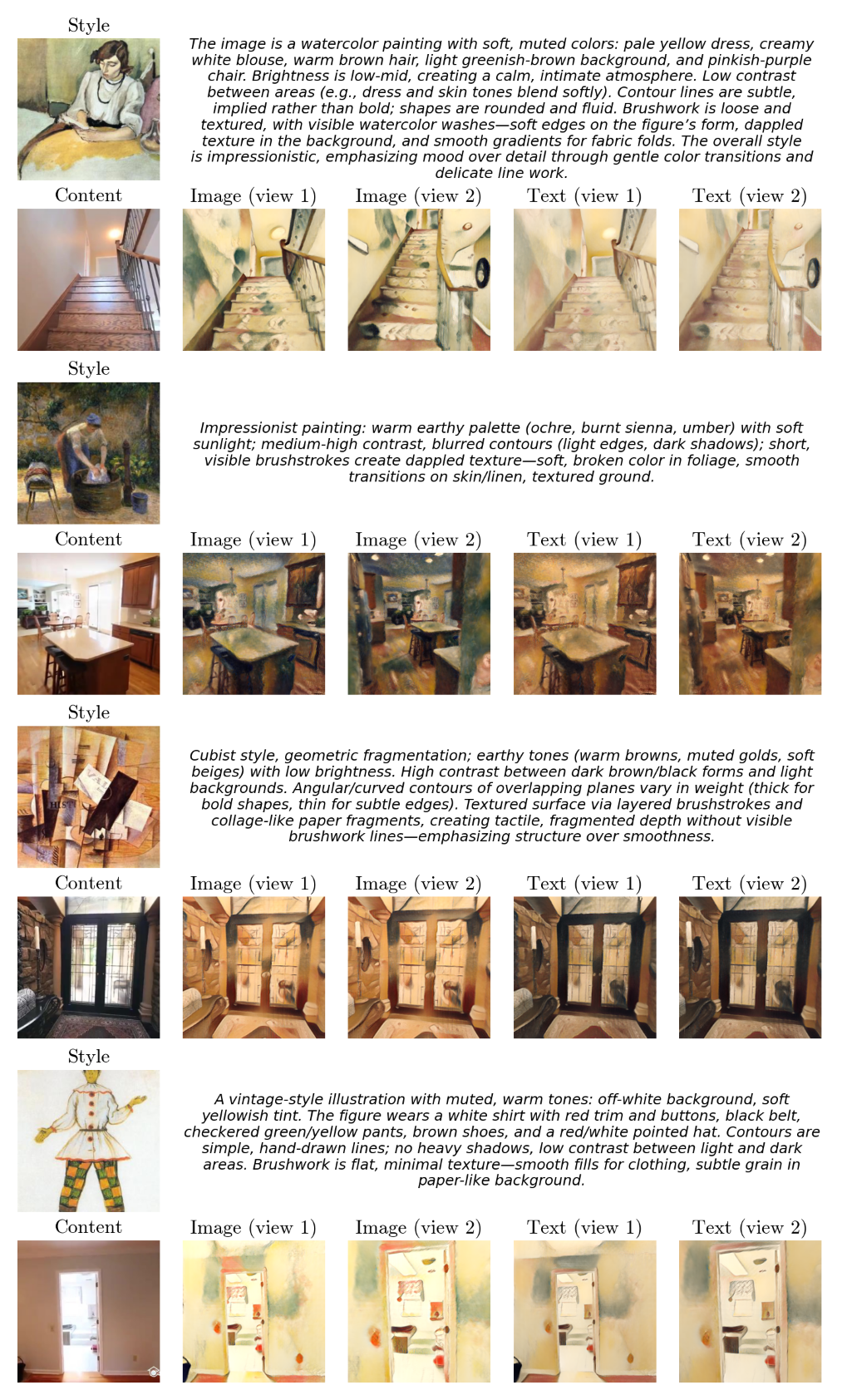}
  \caption{Additional qualitative results of our method on RE10K dataset. We present stylization with image and corresponding text prompt.}
  \label{fig:appx-qual-agg-3}
\end{figure*}

\begin{figure*}
  \centering
  \includegraphics[width=0.75\linewidth]{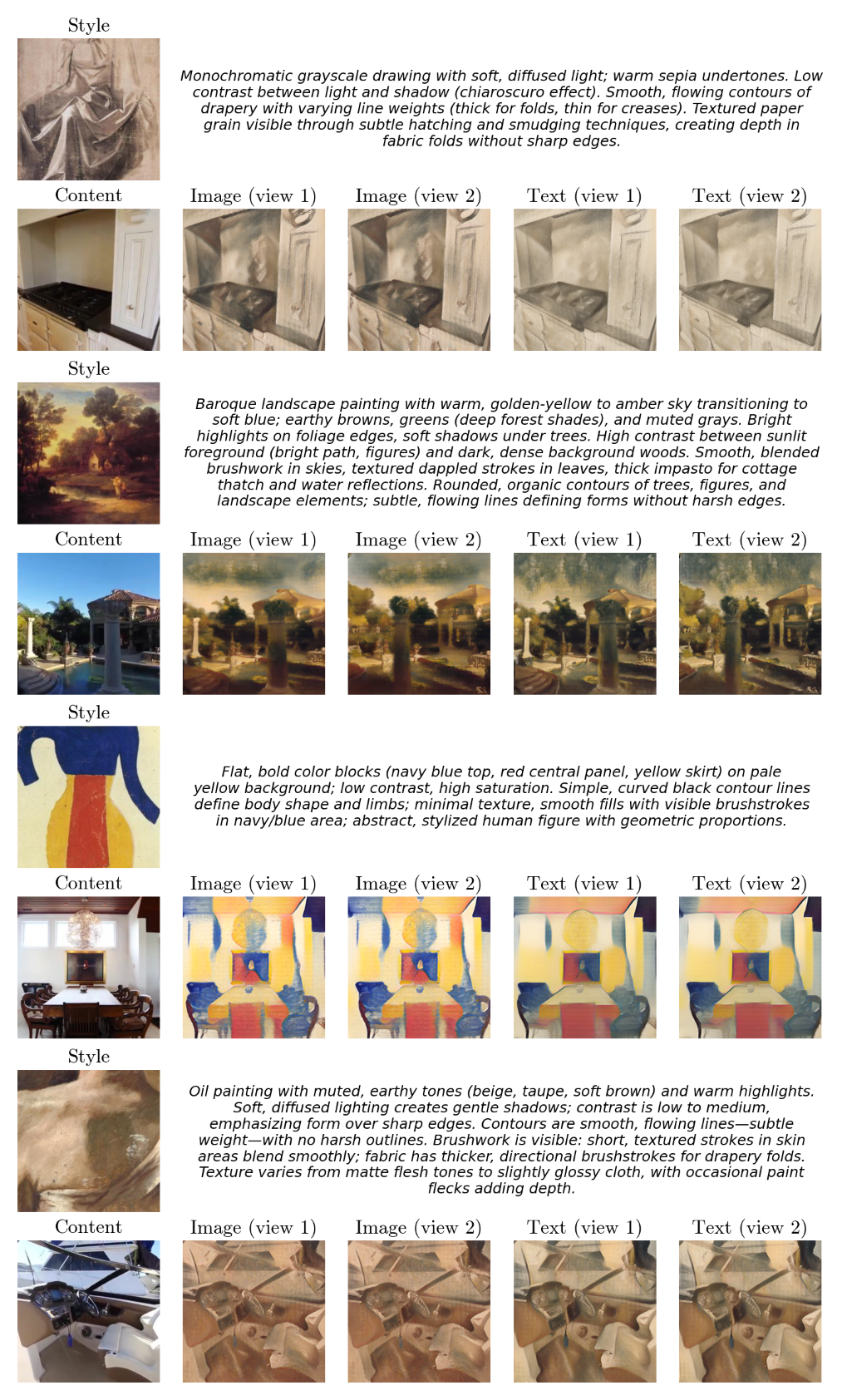}
  \caption{Additional qualitative results of our method on RE10K dataset. We present stylization with image and corresponding text prompt.}
  \label{fig:appx-qual-agg-4}
\end{figure*}

\begin{figure*}
  \centering
  \includegraphics[width=0.75\linewidth]{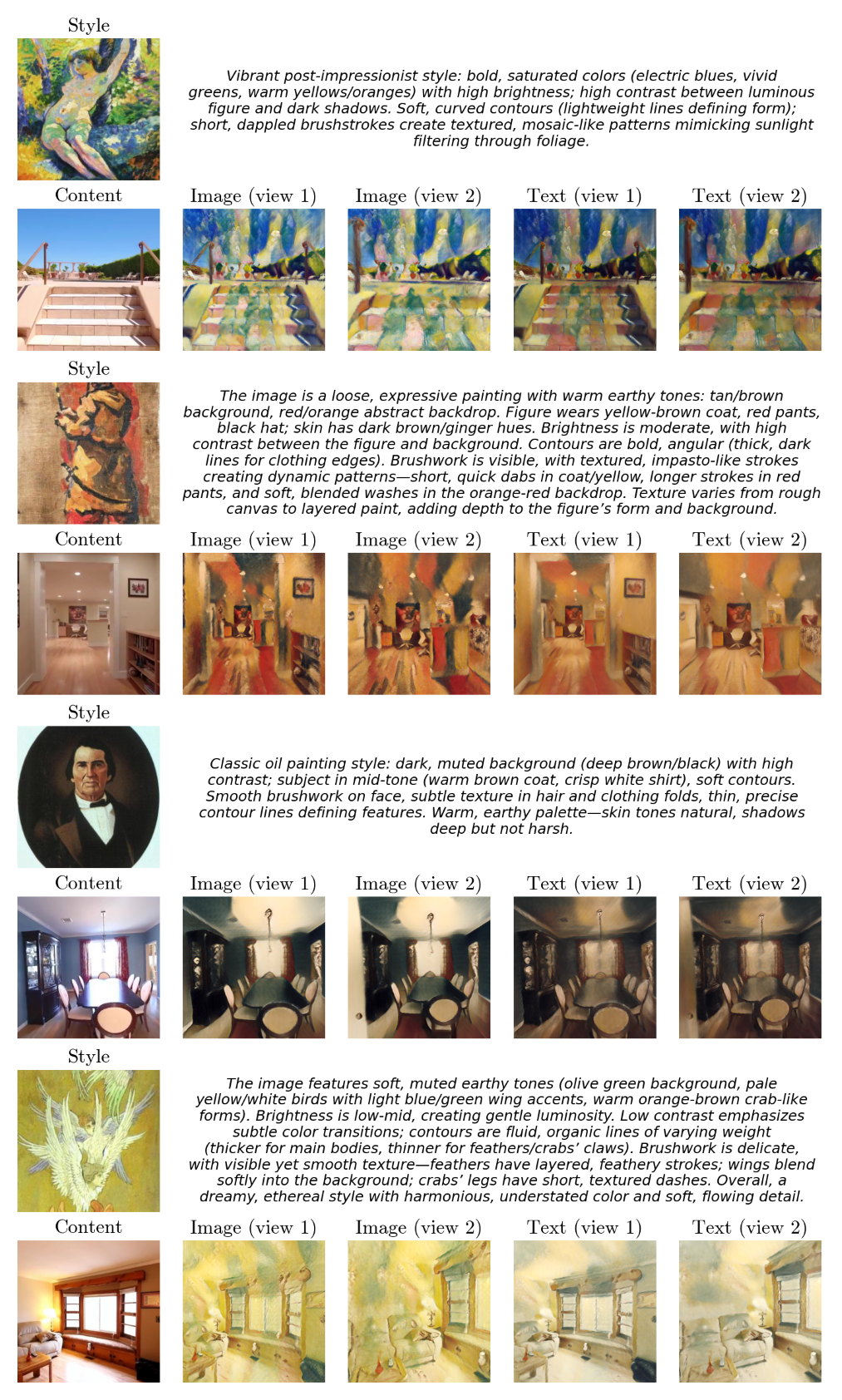}
  \caption{Additional qualitative results of our method on RE10K dataset. We present stylization with image and corresponding text prompt.}
  \label{fig:appx-qual-agg-5}
\end{figure*}

\begin{figure*}
  \centering
  \includegraphics[width=0.90\linewidth]{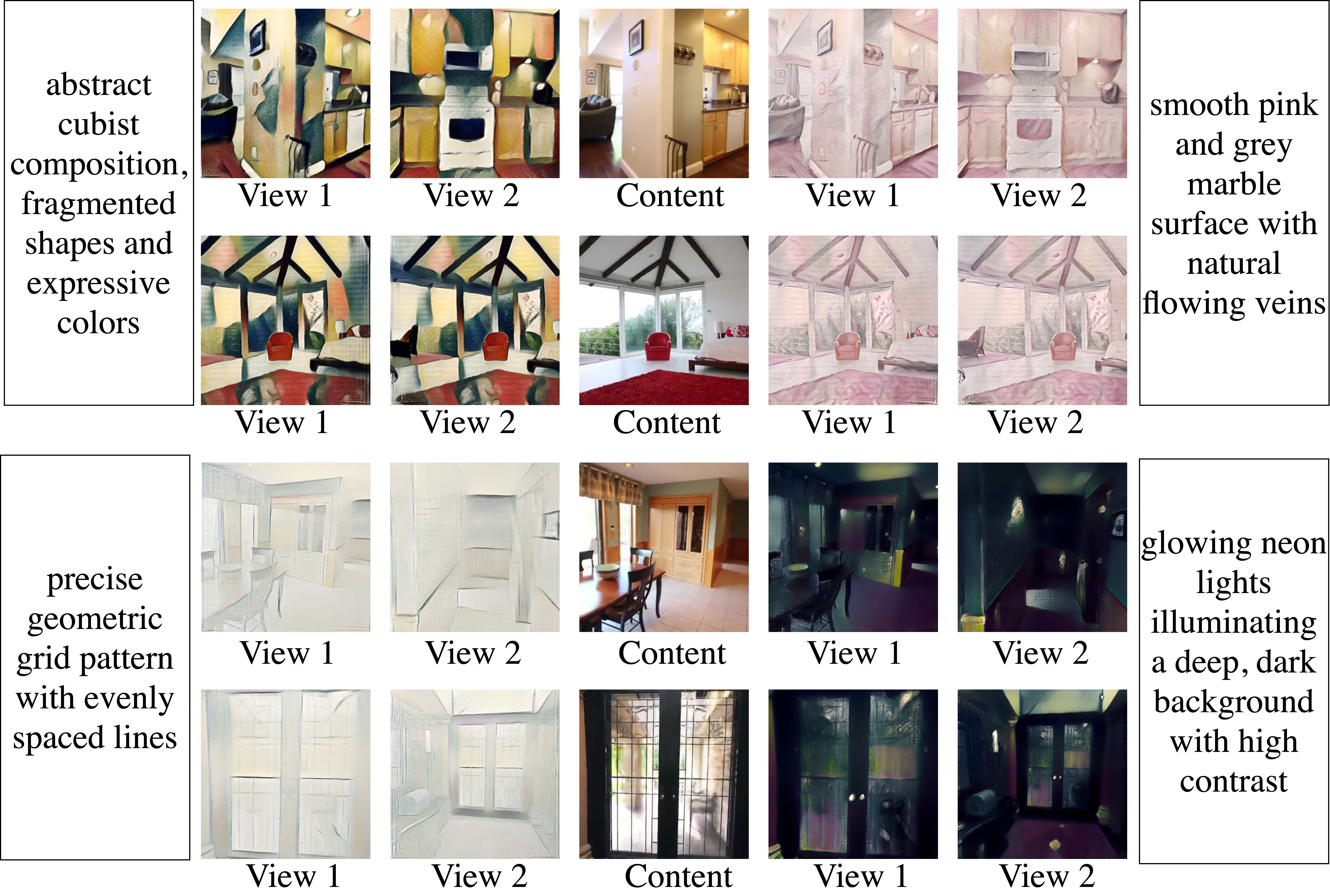}
  \caption{Additional qualitative results of our method on RE10K dataset. We present stylization with natural-language text prompt.}
  \label{fig:appx-natural1}
\end{figure*}

\begin{figure*}
  \centering
  \includegraphics[width=0.90\linewidth]{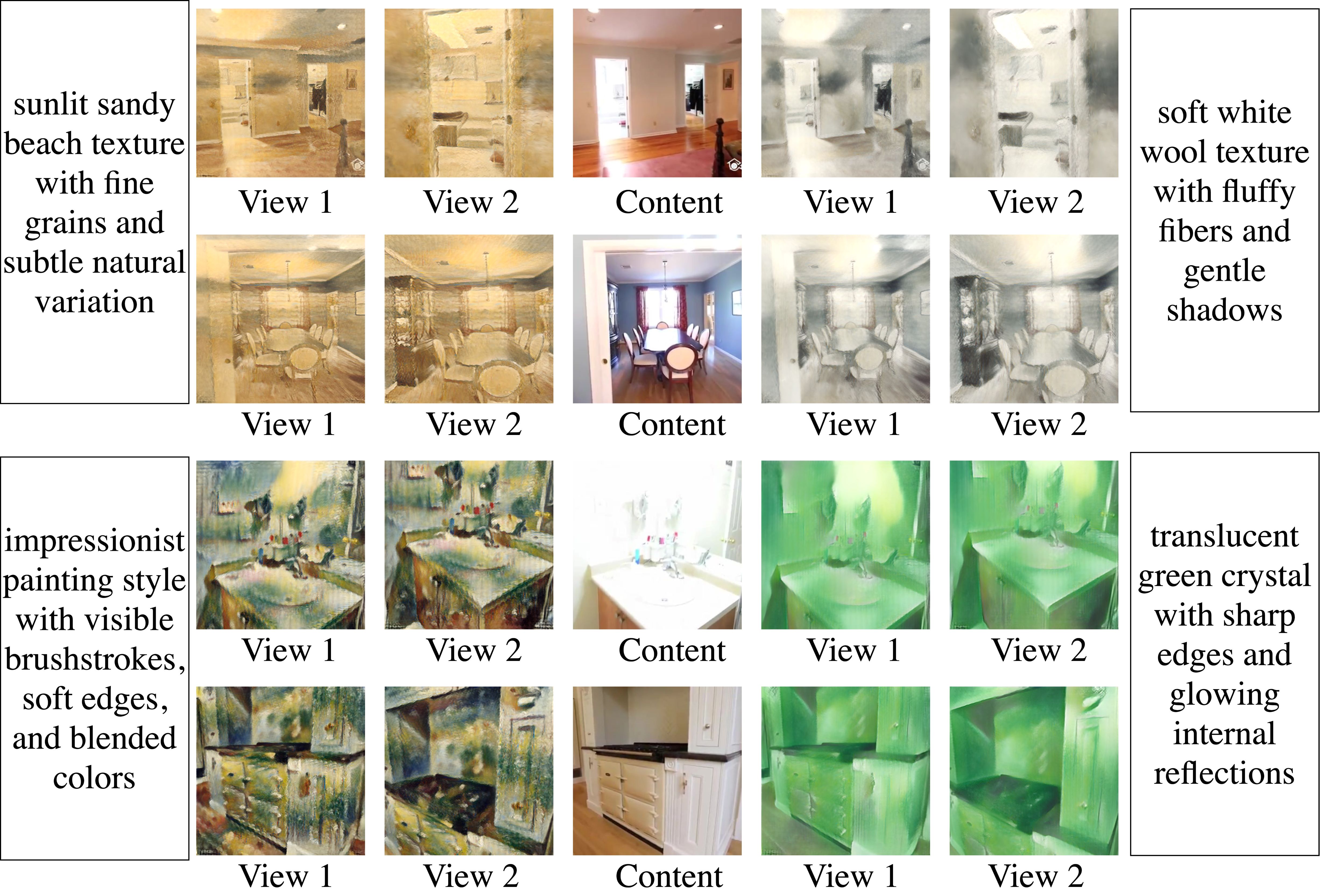}
  \caption{Additional qualitative results of our method on RE10K dataset. We present stylization with natural-language text prompt.}
  \label{fig:appx-natural2}
\end{figure*}

We present additional stylized renders on publicly available RE10K dataset \url{https://google.github.io/realestate10k/}. Text vs image stylization are presented in \cref{fig:appx-qual-agg-1}, \cref{fig:appx-qual-agg-2}, \cref{fig:appx-qual-agg-3}, \cref{fig:appx-qual-agg-4}, \cref{fig:appx-qual-agg-5}, while examples of stylization with natural text are presented in \cref{fig:appx-natural1}, \cref{fig:appx-natural2}.

\clearpage

\section{User study}
\label{sec:user-study-appendix}
 To evaluate the effectiveness of our approach, we conducted a user study involving $40$ participants. Each participant was presented with 20 comparative image sets, resulting in $60$ individual responses per form and a total of $2400$ evaluations across the study. Each set displayed a style reference alongside three reference images and three corresponding generated outputs for two anonymous methods, labeled \textit{Method A} and \textit{Method B}. To eliminate positional bias, the assignment of our method was counterbalanced such that it appeared as \textit{Method A} in half of the trials and \textit{Method B} in the remaining half.

For each comparative image set, participants were asked to answer the following three questions:
\begin{itemize}
    \item On a scale from 1 (A is much better) to 5 (B is much better), rate which method better applied style to images, where 3 indicates methods are comparable.
    \item On a scale from 1 (very difficult) to 5 (very easy), how hard it is to recognize original content on images generated by Method A?
    \item On a scale from 1 (very difficult) to 5 (very easy), how hard it is to recognize original content on images generated by Method B?
\end{itemize}

The results for first question were analyzed using a one-sample T-test and a one-sample Wilcoxon Signed-Rank test with a significance level of $\alpha = 0.05$. For the T-test, the null hypothesis stated that the distribution mean is equal to the population mean (where the neutral value is 3). For the Wilcoxon Signed-Rank test, the null hypothesis was that the distribution of $S-3$ is symmetric around $\mu = 0$, where $S$ is a set of responses. In both cases, the alternative hypothesis was that the distribution mean is smaller, corresponding to the statement that our method performed better than the baseline. We obtained p-values of $3.47 \times 10^{-40}$ and $1.67 \times 10^{-35}$ for the one-sample t-test and the Wilcoxon signed-rank test, respectively. Both values are far below the significance threshold of $\alpha = 0.05$, indicating that the observed preference for our method is statistically significant.

Regarding content preservation, results for both methods were comparable so we have abandoned any statistical test.


\section{Training details}
\label{sec:train-details}
For full implementation and reproduction details, please refer to the supplementary code at \url{https://anonymous.4open.science/r/AnyStyle-13E9/README.md}.

\paragraph{Training sets.} We train our model on the full DL3DV-480P dataset and evaluate on scenes from other datasets. For style supervision we use Wikiart dataset and we exclude 50 test style images from StylOS~\cite{liu2025stylos} as well as the entire Styl3R test set~\cite{wang2025styl3r} to avoid data leakage.

\paragraph{Model training.} Model is trained for 90k iterations. We optimize the copied Gaussian Head and style injectors with a learning rate of $1 \times 10^{-4}$. The Aggregator learning rate is set to $0.3\times$ the original backbone learning rate. We use a cosine annealing learning rate schedule.

\paragraph{Loss weights and parameters.} The training loss consists of a style loss (weight 1.0), a content loss (weight 0.05), a CLIP directional loss (weight 2.0) and patch CLIP loss with 16 crops per image, patch loss weight 4.0 and crop size 64. Depth consistency loss weight is set to 0.1.

\paragraph{Architecture choices.}
Tokens from the intermediate layers $[4, 11, 17, 23]$ are passed to the Gaussian Head, and style injection for Gaussian Head is performed for tokens from these layers. For the Aggregator, features are injected into the tokens before they enter the following layers: $[0, 4, 11, 17, 23]$.

\clearpage
\section{Multi-view consistency: Analysis and Discussion.}
\label{apx:mv-const}
Following most style transfer works~\cite{liu2025stylos, zhang2024stylizedgs}, we measure multi-view consistency. We strictly follow the procedure used in Stylos: we take the first 16 frames and compute a \textit{warp} between two frames using optical flow~\cite{teed2020raft} and softmax splatting~\cite{Niklaus_CVPR_2020}. We then compute masked RMSE and LPIPS~\cite{zhang2018perceptual} on the warped results to evaluate stylization consistency. For \textit{short-range consistency}, we use frame pairs $(t, t\!-\!1)$, and for \textit{long-range consistency}, we use frame pairs $(t, t\!-\!7)$. Results are presented in Table~\ref{tab:consistency}.

\begin{figure}[h!]
  \centering
  \includegraphics[width=0.7\linewidth]{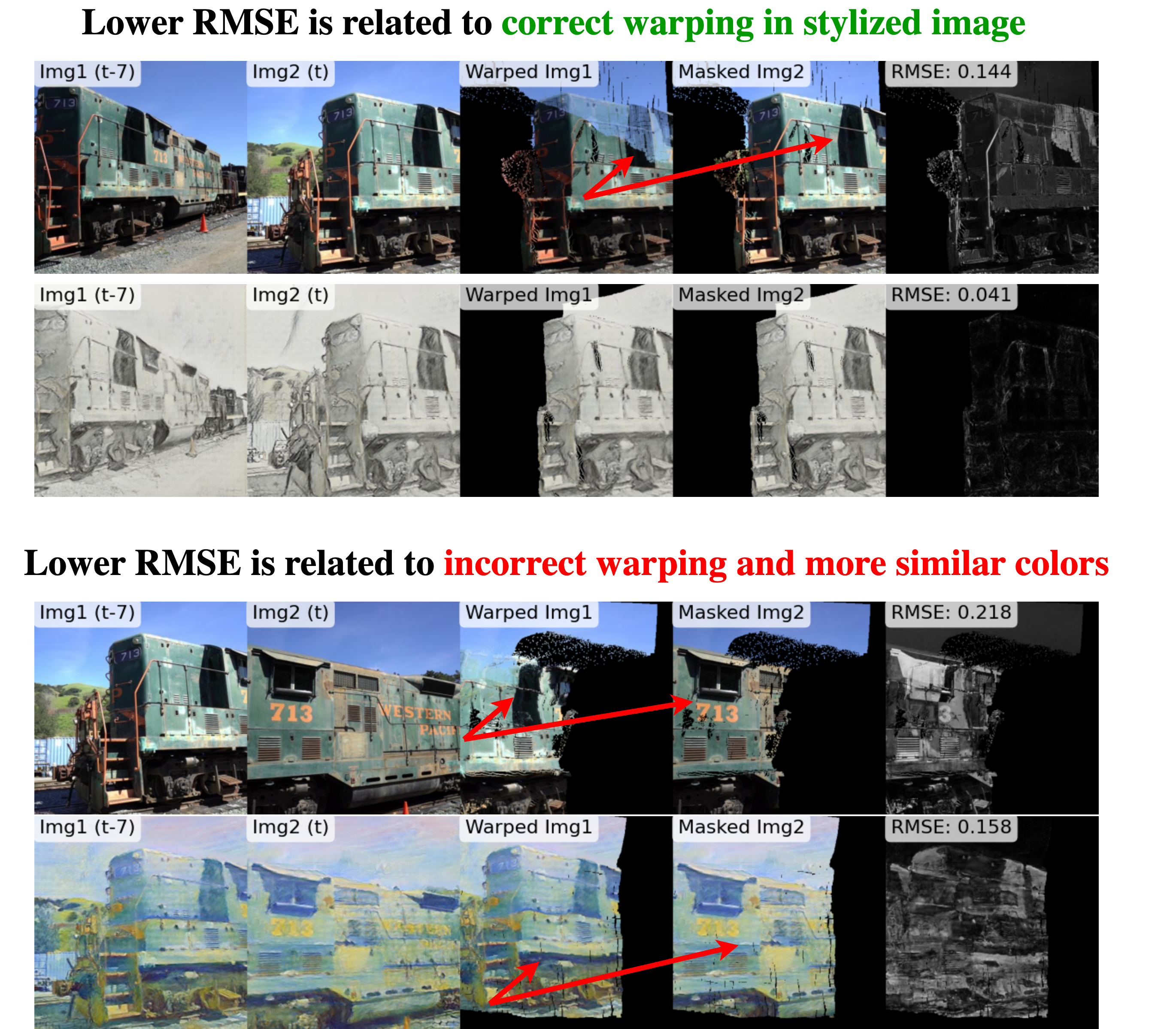}
  \caption{Visualisation of warping and consistency metric RMSE on Train scene. In both sub-figures, stylized images of a train (bottom rows) achieve lower RMSE then GT RGB images (top rows).}
  \label{fig:cons1}
\end{figure}

\begin{figure}
  \centering
  \includegraphics[width=0.7\linewidth]{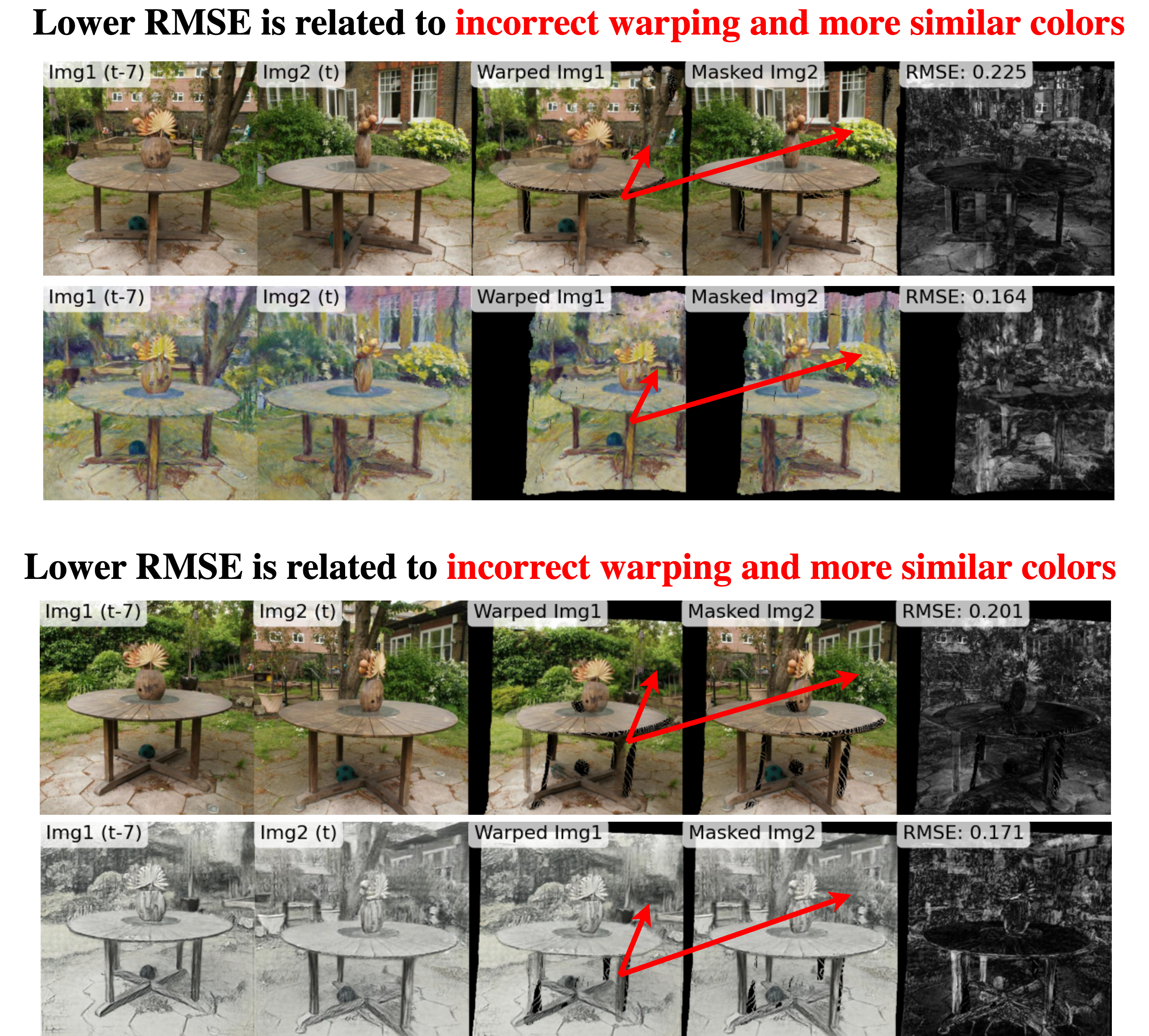}
  \caption{Visualisation of warping and consistency metric RMSE on Garden scene. In both sub-figures, stylized images of a garden (bottom rows) achieve lower RMSE then GT RGB images (top rows).}
  \label{fig:cons2}
\end{figure}

\paragraph{Discussion on the metric.}
We observed that, on the test scenes, the metric performs slighly worse than expected for stylized images and also performs poorly for the \textbf{GT RGB images}. For the GT images, one natural reason we identified is that view-dependent effects can lead to large errors (e.g., specular surface lit by the sun changes appearance with viewpoint). However, we also found a second reason: warping errors, examples of which are shown in \cref{fig:cons1} and \cref{fig:cons2}. We hypothesize that some stylizations can make warping easier (for example, fewer details or more distinct shapes may benefit optical flow, \cref{fig:cons1}). At the same time, we observed that the more uniform colors introduced naturally by stylization can lead to lower RMSE values even when similar warping errors occur in both GT and stylized sequences. In such cases, lower RMSE does not correspond to actual consistency. \textbf{For these reasons, the mean consistency metrics for GT images are sometimes higher (worse) than for our stylized outputs (see \cref{tab:consistency}).} Overall, we conclude that lower metric values may reflect (i) better consistency, (ii) improved warping alignment, or (iii) simply reduced color variation. \textbf{We want to emphasize that these conclusions are valid for both our and baseline methods}. We leave the interpretation to the reader.

\begin{table*}[b!]
\centering
\caption{Short-range and long-range consistency evaluation. Lower is better for all metrics.}
\label{tab:consistency}
\resizebox{\textwidth}{!}{
\begin{tabular}{lcccccccc}
\toprule
\multirow{2}[2]{*}{Method}
& \multicolumn{2}{c}{Train}
& \multicolumn{2}{c}{Truck}
& \multicolumn{2}{c}{M60}
& \multicolumn{2}{c}{Garden} \\
\cmidrule(lr){2-3} \cmidrule(lr){4-5} \cmidrule(lr){6-7} \cmidrule(lr){8-9}
& LPIPS$\downarrow$ & RMSE$\downarrow$
& LPIPS$\downarrow$ & RMSE$\downarrow$
& LPIPS$\downarrow$ & RMSE$\downarrow$
& LPIPS$\downarrow$ & RMSE$\downarrow$ \\
\midrule
\multicolumn{9}{c}{\textit{Short-range consistency}} \\
\midrule
StyleGaussian & 0.033 & 0.038 & 0.031 & 0.034 & 0.038 & 0.037 & 0.069 & 0.061 \\
G-Style & 0.042 & 0.052 & 0.032 & 0.035 & 0.038 & 0.034 & 0.066 & 0.070 \\
StylizedGS & 0.016 & 0.023 & 0.013 & 0.019 & 0.017 & 0.020 & 0.056 & 0.041 \\
SGSST & 0.038 & 0.047 & 0.039 & 0.047 & 0.044 & 0.049 & 0.084 & 0.090 \\
\midrule
Styl3R & 0.056 & 0.038 & 0.049 & 0.033 & 0.064 & 0.042 & 0.085 & 0.061 \\
Stylos & 0.030 & 0.026 & 0.028 & 0.021 & 0.035 & 0.024 & 0.047 & 0.044 \\
\midrule
\our{}$_{Agg, img}$ & 0.032 & 0.038 & 0.033 & 0.034 & 0.042 & 0.041 & 0.063 & 0.063 \\
\our{}$_{Agg, txt}$ & 0.029 & 0.032 & 0.028 & 0.028 & 0.036 & 0.033 & 0.055 & 0.051 \\
\our{}$_{Head, img}$ & 0.032 & 0.035 & 0.031 & 0.031 & 0.041 & 0.036 & 0.066 & 0.060 \\
\our{}$_{Head, txt}$ & 0.029 & 0.030 & 0.028 & 0.026 & 0.038 & 0.030 & 0.059 & 0.049 \\
\midrule
GT images & 0.070 & 0.069 & 0.055 & 0.055 & 0.059 & 0.053 & 0.121 & 0.092 \\
\midrule
\multicolumn{9}{c}{\textit{Long-range consistency}} \\
\midrule
StyleGaussian & 0.067 & 0.072 & 0.086 & 0.077 & 0.091 & 0.091 & 0.177 & 0.141 \\
G-Style & 0.098 & 0.120 & 0.095 & 0.093 & 0.104 & 0.095 & 0.180 & 0.175 \\
StylizedGS & 0.040 & 0.065 & 0.049 & 0.064 & 0.065 & 0.086 & 0.100 & 0.202 \\
SGSST & 0.087 & 0.108 & 0.119 & 0.120 & 0.130 & 0.128 & 0.221 & 0.222 \\
\midrule
Styl3R & 0.109 & 0.090 & 0.136 & 0.112 & 0.160 & 0.138 & 0.185 & 0.171 \\

Stylos & 0.051 & 0.056 & 0.074 & 0.069 & 0.083 & 0.082 & 0.139 & 0.134 \\
\midrule
\our{}$_{Agg, img}$ & 0.082 & 0.088 & 0.092 & 0.094 & 0.115 & 0.117 & 0.198 & 0.170 \\
\our{}$_{Agg, txt}$ & 0.074 & 0.075 & 0.080 & 0.078 & 0.098 & 0.098 & 0.182 & 0.145 \\
\our{}$_{Head, img}$ & 0.070 & 0.073 & 0.091 & 0.089 & 0.115 & 0.108 & 0.189 & 0.147 \\
\our{}$_{Head, txt}$ & 0.062 & 0.061 & 0.079 & 0.074 & 0.103 & 0.091 & 0.169 & 0.122 \\
\midrule
GT images & 0.130 & 0.142 & 0.129 & 0.113 & 0.147 & 0.123 & 0.284 & 0.206 \\

\bottomrule
\end{tabular}
}
\vspace{0mm}
\end{table*}

\clearpage

\section{Extended quantitative results. }
\label{sec:ext-quant}
In \cref{tab:more_metrics_1,tab:more_metrics_2}, following \cite{styleid, liu2025stylos} we provide extended quantitative evaluation of stylized data. Notably, we \textbf{consistently achieve strong (low) FID, LPIPS and LPIPS-Gray results}. 
\begin{table*}[h!]
\centering
\caption{Following Stylos~\cite{liu2025stylos} and StyleID \cite{styleid}, we additionally report stylization quality metrics: FID, LPIPS, LPIPS-gray, CFSD, and color matching loss (HistoGAN loss), as supplementary to results in main paper.}
\label{tab:more_metrics_1}
\begin{tabular}{lcccccccccc}
\toprule
\multirow{2}[2]{*}{Method} & \multicolumn{5}{c}{Train} & \multicolumn{5}{c}{Truck} \\
\cmidrule(lr){2-6}\cmidrule(lr){7-11}
 & FID$\downarrow$ & LPIPS$\downarrow$ & Gray$\downarrow$ & CFSD$\downarrow$ & CM Loss$\downarrow$
       & FID$\downarrow$ & LPIPS$\downarrow$ & Gray$\downarrow$ & CFSD$\downarrow$ & CM Loss$\downarrow$ \\
\midrule
StyleGaussian & 34.59 & 0.483 & 0.377 & 0.190 & 0.418 & 28.53 & 0.522 & 0.367 & 0.131 & 0.382 \\
G-Style       & 14.80 & 0.471 & 0.364 & 0.165 & 0.185 & 13.65 & 0.512 & 0.353 & 0.087 & 0.176 \\
StylizedGS    & 22.90 & 0.707 & 0.648 & 0.216 & 0.396 & 22.65 & 0.779 & 0.713 & 0.084 & 0.417 \\
SGSST         & 24.15 & 0.520 & 0.409 & 0.220 & 0.257 & 19.50 & 0.577 & 0.445 & 0.177 & 0.196 \\
Styl3R        & 19.77 & 0.670 & 0.598 & 0.244 & 0.364 & 19.28 & 0.682 & 0.575 & 0.102 & 0.350 \\
Stylos        & 15.30 & 0.620 & 0.529 & 0.223 & 0.241 & 16.67 & 0.625 & 0.471 & 0.084 & 0.237 \\
\midrule
\our{}$_{Agg, img}$   & 13.81 & 0.543 & 0.457 & 0.256 & 0.359 & 13.61 & 0.570 & 0.433 & 0.112 & 0.377 \\
\our{}$_{Agg, txt}$    & 15.08 & 0.518 & 0.421 & 0.236 & 0.438 & 14.97 & 0.544 & 0.399 & 0.090 & 0.447 \\
\our{}$_{Head, img}$    & 13.73 & 0.551 & 0.452 & 0.249 & 0.364 & 14.94 & 0.586 & 0.437 & 0.103 & 0.390 \\
\our{}$_{Head, txt}$    & 14.80 & 0.529 & 0.421 & 0.238 & 0.442 & 16.74 & 0.572 & 0.418 & 0.091 & 0.465 \\
\bottomrule
\end{tabular}
\end{table*}

\begin{table*}[h!]
\centering
\caption{Following Stylos~\cite{liu2025stylos} and StyleID \cite{styleid}, we additionally report stylization quality metrics: FID, LPIPS, LPIPS-gray, CFSD, and color matching loss (HistoGAN loss), as supplementary to results in main paper.}
\label{tab:more_metrics_2}
\begin{tabular}{lcccccccccc}
\toprule
\multirow{2}[2]{*}{Method} & \multicolumn{5}{c}{M60} & \multicolumn{5}{c}{Garden} \\
\cmidrule(lr){2-6}\cmidrule(lr){7-11}
 & FID$\downarrow$ & LPIPS$\downarrow$ & Gray$\downarrow$ & CFSD$\downarrow$ & CM Loss$\downarrow$
       & FID$\downarrow$ & LPIPS$\downarrow$ & Gray$\downarrow$ & CFSD$\downarrow$ & CM Loss$\downarrow$ \\
\midrule
StyleGaussian & 30.54 & 0.506 & 0.413 & 0.138 & 0.467 & 25.23 & 0.569 & 0.454 & 0.189 & 0.480 \\
G-Style       & 13.93 & 0.498 & 0.395 & 0.092 & 0.208 & 16.17 & 0.500 & 0.364 & 0.103 & 0.179 \\
StylizedGS    & 28.33 & 0.815 & 0.728 & 0.102 & 0.443 & 33.73 & 0.876 & 0.827 & 0.074 & 0.570 \\
SGSST         & 23.95 & 0.552 & 0.458 & 0.204 & 0.264 & 20.93 & 0.529 & 0.415 & 0.233 & 0.228 \\
Styl3R        & 17.14 & 0.646 & 0.573 & 0.124 & 0.314 & 22.38 & 0.637 & 0.556 & 0.097 & 0.335 \\
Stylos        & 16.61 & 0.558 & 0.457 & 0.098 & 0.252 & 16.26 & 0.625 & 0.509 & 0.080 & 0.242 \\
\midrule
\our{}$_{Agg, img}$        & 13.88 & 0.533 & 0.456 & 0.131 & 0.373 & 13.43 & 0.546 & 0.467 & 0.114 & 0.375 \\
\our{}$_{Agg, txt}$        & 15.07 & 0.503 & 0.429 & 0.111 & 0.450 & 14.76 & 0.536 & 0.447 & 0.086 & 0.446 \\
\our{}$_{Head, img}$        & 13.79 & 0.543 & 0.452 & 0.120 & 0.369 & 14.44 & 0.545 & 0.452 & 0.100 & 0.380 \\
\our{}$_{Head, txt}$        & 15.00 & 0.521 & 0.427 & 0.108 & 0.452 & 15.60 & 0.539 & 0.440 & 0.085 & 0.465 \\

\bottomrule
\end{tabular}
\end{table*}

\clearpage

\section{Limitations \& failure cases}
\label{sec:limitations}
\begin{figure}[h]
  \centering
  \includegraphics[width=0.8\linewidth]{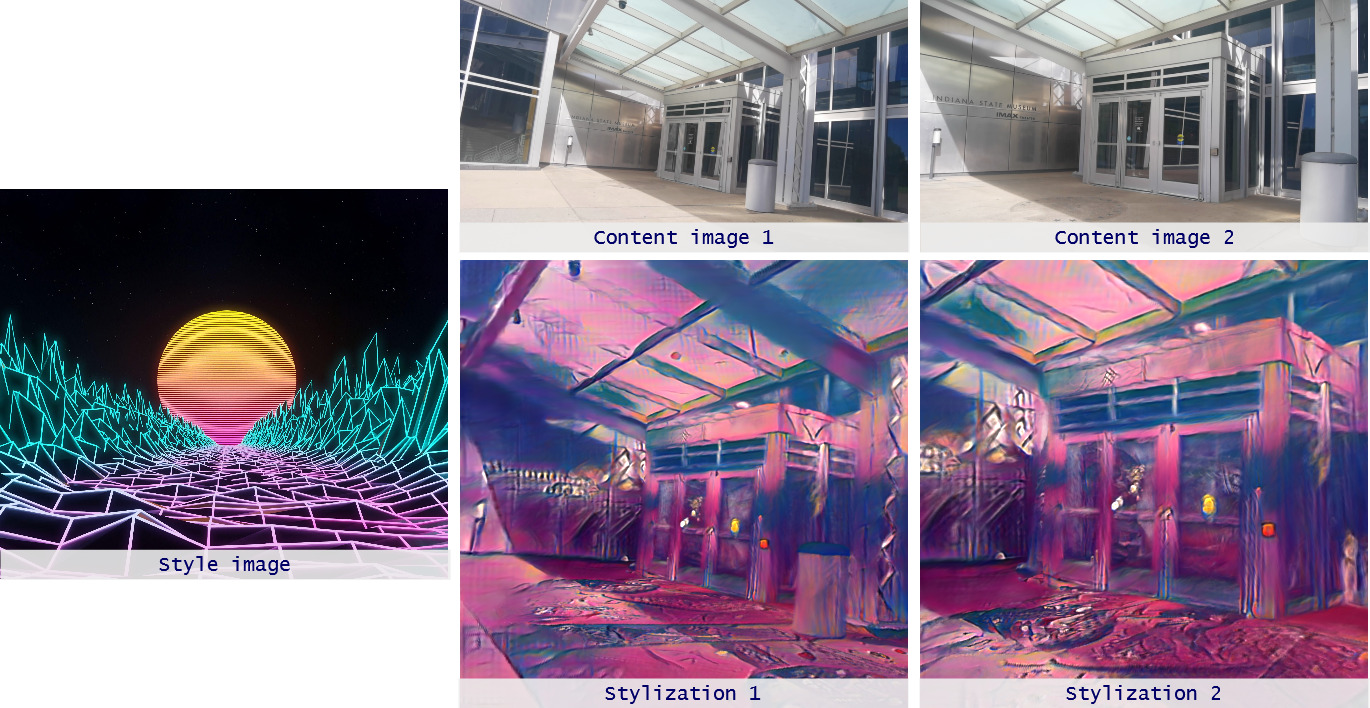}
  \caption{Example failure case when providing image as conditioning signal. Model fails to transfer the essence of the style. Instead it applies only color pallette from style image.}
  \label{fig:failure1}
\end{figure}

\begin{figure}[h]
  \centering
  \includegraphics[width=0.8\linewidth]{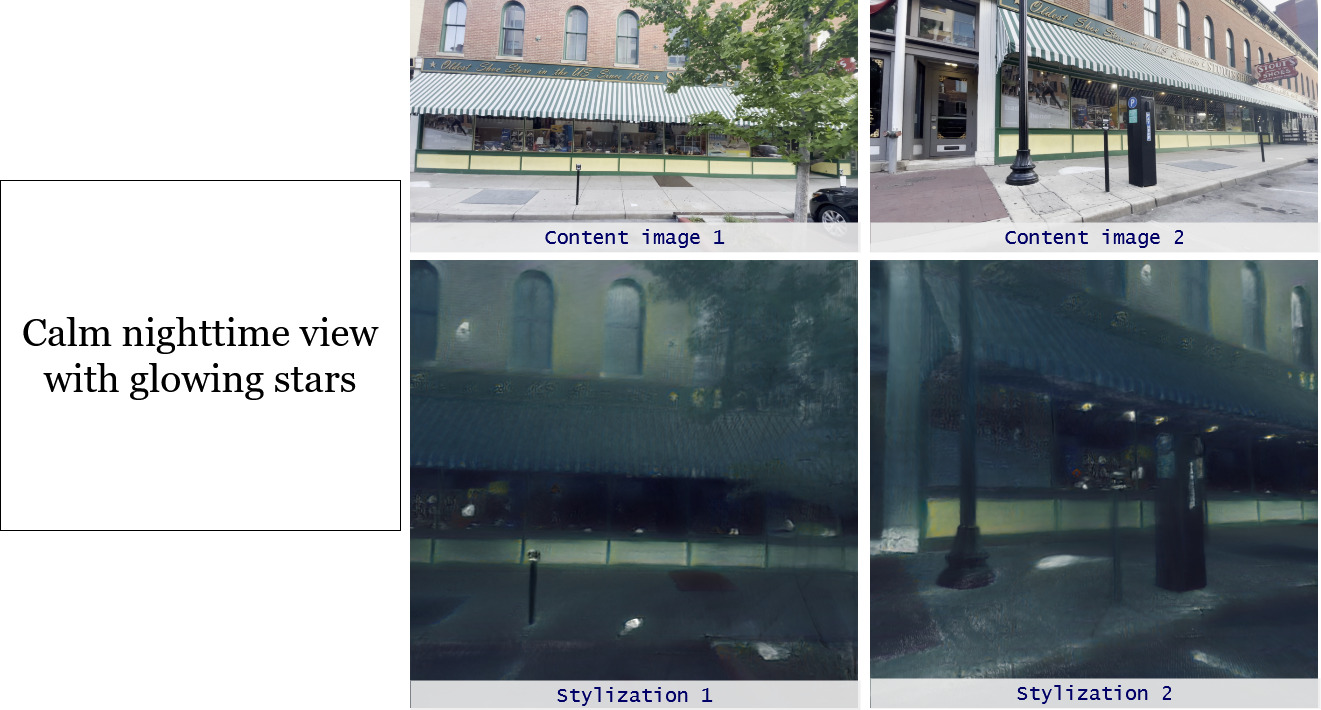}
  \caption{Example failure case when providing text as conditioning signal. While overall aesthetics my be pleasing, model fails to create realistic nighttime view.}
  \label{fig:failure2}
\end{figure}

While our method performs well across a wide range of styles for both image and text stylization, its performance may degrade for inputs that are far outside the WikiArt distribution, where our train artistic style examples originate. In such out-of-distribution cases, the stylization can present some visual properties of target style, but can be less visually appealing. ~\cref{fig:failure1}, \cref{fig:failure2} show representative failure cases, including an extreme example with neon-like line structures and a non-artistic text prompt. A potential direction to reduce these failures - especially in strongly out-of-distribution cases is test-time embedding optimization.

\end{document}